%% file: draft.tex
\documentclass{article}
\usepackage[dvipsnames]{xcolor}

\usepackage{titling}  
\usepackage{float}
\usepackage{longtable}
\usepackage{adjustbox, textgreek}
\usepackage{pdflscape}
\usepackage{multirow}
\usepackage{multicol}
\usepackage{makecell}
\usepackage[most]{tcolorbox}
\usepackage{booktabs}
\usepackage{array}
\usepackage{rotating}
\usepackage{bbding}
\usepackage[edges]{forest}
\usepackage{makecell}
\usepackage{graphicx}
\usepackage{multirow}
\usepackage{caption}
\usepackage{subcaption}
\usepackage{adjustbox}
\usepackage{amsmath}
\usepackage{ltablex}
\usepackage{tikz}
\usepackage{titlesec}
\usepackage{titling}
\usetikzlibrary{trees,positioning,shapes,shadows,arrows.meta}
\usepackage[
labelfont=sf,
hypcap=false,
format=hang,
width=\columnwidth
]{caption}

\definecolor{light_red}{rgb}{1, 0.5, 0.5}
\definecolor{light_blue}{rgb}{0.5, 0.5, 1}
\definecolor{light_orange}{rgb}{1, 0.75, 0.5}
\definecolor{dark_green}{rgb}{0, 0.42, 0.3}

\newcommand*\samethanks[1][\value{footnote}]{\footnotemark[#1]}
\usepackage{bbm}
\usepackage{authblk}

\input{structure.tex} 
\input{math}
\pretitle{%
  \noindent\rule{\textwidth}{1pt}\par
  \begin{center}\LARGE\bfseries
}
\posttitle{%
  \end{center}
  \noindent\rule{\textwidth}{1pt}\par
}

\usepackage{fancyhdr}
\usepackage{etoolbox}  

\usepackage{fancyhdr}
\makeatletter
\patchcmd{\@maketitle}{\thispagestyle{plain}}{\thispagestyle{fancy}}{}{}
\makeatother

\makeatletter
\renewcommand\AB@affilsepx{, \protect\Affilfont}
\makeatother

\renewcommand\Affilfont{\itshape\normalsize}

\title{Reasoning on a Budget: A Survey of Adaptive and Controllable Test-Time Compute in LLMs}

\author[1]{Mohammad Ali Alomrani \thanks{Corresponding authors: \{mohammad.ali.alomrani, yingxue.zhang\}@huawei.com}}
\author[1]{Yingxue Zhang\samethanks}
\author[1]{Derek Li}
\author[1]{Qianyi Sun}
\author[1]{Soumyasundar Pal}
\author[1]{Zhanguang Zhang}
\author[1]{Yaochen Hu}
\author[1]{Rohan Deepak Ajwani}
\author[2 3]{Antonios Valkanas}

\author[1]{Raika Karimi}
\author[1]{Peng Cheng}
\author[1]{Yunzhou Wang}
\author[1]{Pengyi Liao}
\author[1]{Hanrui Huang}
\author[1]{Bin Wang}
\author[1]{Jianye Hao}
\author[2 3]{Mark Coates}
\affil[1]{Huawei Noah's Ark Lab}
\affil[2]{McGill University}
\affil[3]{Mila-Quebec AI Institute}

\date{} 

\begin{document}

\maketitle 

\begin{abstract}
\normalsize
Large language models (LLMs) have rapidly progressed into general-purpose agents capable of solving a broad spectrum of tasks. However, current models remain inefficient at reasoning: they apply fixed inference-time compute regardless of task complexity, often overthinking simple problems while underthinking hard ones. This survey presents a comprehensive review of efficient test-time compute (TTC) strategies, which aim to improve the computational efficiency of LLM reasoning. We introduce a two-tiered taxonomy that distinguishes between L1 \textit{controllability}—methods that operate under fixed compute budgets—and L2 \textit{adaptiveness}—methods that dynamically scale inference based on input difficulty or model confidence. We benchmark leading proprietary LLMs across diverse datasets, highlighting critical trade-offs between reasoning performance and token usage. Compared to prior surveys on efficient reasoning, our review emphasizes the practical control, adaptability, and scalability of TTC methods. Finally, we discuss emerging trends such as hybrid thinking models and identify key challenges for future work towards making LLMs more computationally efficient, robust, and responsive to user constraints.
\end{abstract}

\section{Introduction}
In just a few years, large language models (LLMs) have achieved remarkable success as general-purpose agents, capable of handling tasks ranging from coding~\citep{jiang2024surveylargelanguagemodels} and mathematical reasoning~\citep{ahn2024LargeLanguageModels} to image generation~\citep{openai2025gpt4oimage} and mobile robotic manipulation~\citep{intelligence2025pi05visionlanguage}. This success can be attributed to three core ingredients. First, LLMs are largely built on the transformer architecture~\citep{vaswani2017attentionallneed},  which supports parallelization and handles diverse data modalities~\citep{xu2023multimodalLearning}. Second, by representing all tasks and modalities as sequences of discrete tokens, LLMs can perform a wide variety of tasks with minimal task-specific customization. Users can simply prompt the model with a problem description and a few examples~\citep{brown2020LanguageModelsfewshot}. Third, neural scaling laws~\citep{kaplan2020scalinglawsneurall} demonstrate that increasing model size, data, and compute results in predictable gains in performance. Together, these insights have driven the development of today’s billion- and trillion-parameter models~\citep{meta2025llama4}, which continue to achieve state-of-the-art results on many benchmarks.

Despite these advances, LLMs still struggle with complex reasoning, planning, and search tasks, such as spatial~\citep{yamada2024evaluatingSpatialUnderstanding} and embodied reasoning~\citep{valmeekam2023onplanningabilities}. Unlike humans, who can flexibly allocate cognitive effort when solving difficult problems~\citep{kahneman2012thinkingfastslow}, LLMs typically use a fixed amount of computation (i.e., fixed number of layers) per token during inference, regardless of task difficulty. To address this limitation, a growing body of research has focused on test-time compute (TTC) methods~\citep{wang2023selfconsistency, li2025fastmctssimples, openai2024openaio1card, deepseekai2025deepseekr1}, which aim to improve LLM reasoning by allocating additional computation at inference time. These methods allow LLMs to decompose complex problems into intermediate steps~\citep{wei2022chainofthought} or explore multiple solution paths in parallel~\citep{brown2024largelanguagemonkeys, snell2024scaling}.

While TTC methods have led to significant gains in reasoning-heavy domains such as mathematics and planning~\citep{huang2025mathperturb}, they can be highly inefficient—often consuming over 10× the compute of a standard forward pass~\citep{brown2024largelanguagemonkeys}. Large reasoning models~\citep{deepseekai2025deepseekr1, openai2024openaio1card, team2025kimi} frequently waste resources by ``overthinking'' simple queries (e.g., ``1+1=?'') while failing to allocate sufficient compute to more complex problems~\citep{chen2025think23overthinking}. This inefficiency poses challenges for deploying TTC methods in real-world, time-sensitive applications such as autonomous driving, robotics, and virtual assistants. As a result, recent research~\citep{xia2025tokenskipcontrollable, lee2025llmscompresschainofthought, xu2025chainofdraft, aggarwal2023adaptive, aggarwal2025l1controllinglongreasoning} has turned toward more efficient TTC strategies—that is, methods that adapt or constrain inference-time computation based on user preferences or task difficulty.

The growing demand for practical, real-time applications has also led commercial LLM developers to invest heavily in controllable and efficient TTC mechanisms. Leading AI labs are increasingly adopting the concept of {\em fast-slow} thinking, where models can modulate their reasoning depth based on the complexity of the task. For example, Anthropic’s Claude 3.7~\citep{anthropic2024claude37systemcard} series supports a ``thinking token budget'' that allows users to trade off speed for deeper reasoning. Similarly, OpenAI’s o1 variants~\citep{openai2024openaio1card, openai2025o3} offer models with several {\em thinking modes} to support different latency and cost requirements. These offerings reflect a growing recognition that one-size-fits-all inference is suboptimal—especially in scenarios where compute cost, latency, and accuracy must be balanced dynamically. This emerging product trend highlights the practical significance of research on controllable and adaptive test-time compute, and underscores the urgent need for principled methods that enable LLMs to reason more efficiently under user-defined constraints.

\begin{figure}[t]
    \centering

\tikzset{
    basic/.style  = {rectangle,
    draw,
    rounded corners,
    text opacity=1,
    minimum height=1.5em,
    minimum width=5em,
    inner sep=2pt,
    align=center,
    fill opacity=.5,},
    root/.style   = {basic, rounded corners=2pt, thin, align=center,child anchor=north, parent anchor=south, anchor=center},
    onode/.style  = {basic, thin, rounded corners=2pt, align=center, text width=3cm},
    tnode1/.style  = {basic, thin, align=center, fill=brown!30, text width=15em},
    tnode2/.style  = {basic, thin, align=center, fill=cyan!10, text width=15em},
    tnode3/.style  = {basic, thin, align=center, fill=gray!10, text width=15em},
    tnode4/.style  = {basic, thin, align=center, fill=yellow!20, text width=15em},
    xnode/.style  = {basic, thin, rounded corners=2pt, align=center, text width=3cm},
    xnode1/.style  = {basic, thin, rounded corners=2pt, align=center, text width=2cm},
    edge from parent/.style={draw=black, edge from parent fork right}
}
\scalebox{0.7}{
\begin{forest}
  forked edges,
    for tree={
      grow=east,
       growth parent anchor=west,
       parent anchor=east,
       child anchor=west,
       minimum width=6em,
    edge path={
      \noexpand\path[\forestoption{edge},-,>=latex]
        (!u.parent anchor) -- +(10pt,0pt) |- (.child anchor)
        \forestoption{edge label};
    }
    },
  [\textbf{Precise and Efficient}\\\textbf{Test-time Compute}\\\textbf{for LLMs}, root, l sep=10mm
    [\textbf{L1: Controllable} \\ (\S~\ref{sec:l1}), xnode, l sep=10mm
      [\textbf{Parallel}\\ (\S~\ref{sec:l1_parallel}), xnode1, l sep=10mm
        [\textbf{Prompting-based} \\ 
        Self-Consistancy~\citet{wang2023selfconsistency}{,}
        BoN~\citet{brown2024largelanguagemonkeys}{,} Beam Search~\citet{kang2024mindstarenhancing}{,} 
        Lookahead Search~\citet{snell2024scaling}{,}
        ToT~\citet{yao2023treeofthought}{,} MCTS~\citet{li2025fastmctssimples}, tnode1, text width=40em]
        [\textbf{Supervised Finetuning (SFT)} \\  BoN-SFT~\citet{chow2025inferenceawarefinetuning}{,} , tnode1, text width=40em]
        [\textbf{Reinforcement Learning (RL)} \\ BoN-RL~\citet{chow2025inferenceawarefinetuning}{,}, tnode1, text width=40em]
      ]
      [\textbf{Sequential} \\ (\S~\ref{sec:l1_sequential}), xnode1, l sep=10mm
        [\textbf{Prompting-based} \\ 
        CCoT~\citet{renze2024BenefitsConciseChain}\text{,} 
        CCoT~\citet{nayab2025concisethoughtsimpact}\text{,} 
        CoD~\citet{xu2025chainofdraft}\text{,}
        ToT~\citet{yao2023treeofthought}\text{,} 
        Beam Search~\citet{kang2024mindstarenhancing}{,}
        MCTS~\citet{li2025fastmctssimples}{,}
        s1~\citet{muennighoff2025s1simpletest}, tnode2, text width=40em]
        [\textbf{Supervised Finetuning (SFT)} \\ CoT-Valve~\citet{ma2025cotvalvelengthcompressible}\text{,} Sequential Revision~\citet{snell2024scaling}{,}
        TokenSkip~\citet{xia2025tokenskipcontrollable}\text{,} System-1.x~\citet{saha2025system1xlearning}, tnode2, text width=40em]
        [\textbf{Reinforcement Learning (RL)} \\ L1~\citet{aggarwal2025l1controllinglongreasoning}{,}
        Elastic Reasoning~\citet{xu2025scalablechainthoughtselastic}{,}
        SCoRe~\citet{kumar2024traininglanguagemodels}{,} o1~\citet{openai2024openaio1card}{,} Claude 3.7 Sonnet~\citet{anthropic2024claude37systemcard}{,} Grok 3~\citet{xai2025grok3}, tnode2, text width=40em]
      ]
    ]
    [\textbf{L2: Adaptive} \\ (\S~\ref{sec:l2}), xnode, l sep=10mm
      [\textbf{Parallel} \\ (\S~\ref{sec:l2_parallel}), xnode1, l sep=10mm
        [\textbf{Prompting-based} \\ Adaptive-Consistency~\citet{aggarwal2023adaptive}{,} ESC~\citet{li2024escape}{,} Path-Consistency~\citet{zhu2024path_consistency}{,} DSC~\citet{wang2024make}{,} DynaThink~\citet{pan2024dynathink}{,} Dynasor~\citet{fu2024certainindex}{,} Length-filtered Vote~\citet{wu2025whenmoreless}{,} DPTS~\citet{ding2025dynamicparalleltree}{,} ST-BoN~\citet{wang2025self-estimating}{,} FastMCTS~\citet{li2025fastmctssimples}{,} Speculative Rejection~\citet{sun2024fastbestofn}{,} Analytical Scaling Model~\citet{chen2024aremorellm}{,} etc, tnode3, text width=40em]
        [\textbf{Supervised Finetuning (SFT)} \\ RASC~\citet{wan2024rasc}{,} Adaptive Allocation~\citet{manvi2024}{,} Ada-BoK~\citet{damani2025learning}{,} Self-Calibration~\citet{huang2025tts-self-calibrate}{,} HaluSearch~\citet{cheng2025thinkmorehallucinate}{,} etc, tnode3, text width=40em]
        [\textbf{Reinforcement Learning (RL)} \\ Adaptive Parallel Reasoning~\citet{pan2025learningadaptiveparallel}, tnode3, text width=40em]
      ]
      [\textbf{Sequential} \\ (\S~\ref{sec:l2_sequential}), xnode1, l sep=10mm
        [\textbf{Prompting-based} \\ SoT~\citet{aytes2025sketchofthought}\text{,} TIP~\citet{wang2025thoughtsplacethinking}\text{,} CCoT~\citet{renze2024BenefitsConciseChain}\text{,}MetaReasoner~\citet{sui2025metareasonerdynamic}, tnode4, text width=40em]
        [\textbf{Supervised Finetuning (SFT)} \\ 
        Pangu Embedded~\citet{chen2025panguembeddedefficient}{,}
        TOPS~\citet{yang2025thinkingoptimalscaling}\text{,} TALE~\citet{han2025tokenbudgetaware}\text{,} 
        MRT~\citet{qu2025optimizing}{,}
        DualFormer~\citet{su2025dualformerControllableFast}\text{,} C3oT~\citet{kang2024c3otgeneratingshorterchain}\text{,} Self-training~\citet{munkhbat2025selftrainingelicits}\text{,} Z1~\citet{yu2025z1efficienttesttimesscaling}\text{,} Distillation~\citet{yu2024distilling21}\text{,} Retro-Search~\citet{lu2025retrosearchexploring}\text{,} SoftCoT~\citet{xu2025softcotsoftchain}{,} Coconut~\citet{hao2024traininglargelanguage}{,} CODI~\citet{shen2025codicompressingchain}{,} CCoT~\citet{shen2025efficientreasoninghidden}{,} etc, tnode4, text width=40em]
        [\textbf{Reinforcement Learning (RL)} \\ Kimi~\citet{team2025kimi}{,} Length Preference Optimization~\citet{chen2025think23overthinking}{,} O1-pruner~\citet{luo2025o1}{,} DAST~\citet{shen2025dast}{,} MRT~\citet{qu2025optimizing}{,} CPO~\citet{zhang2024chain}{,} IBPO~\citet{yu2025think}{,} etc, tnode4, text width=40em]
      ]
    ]
  ]
\end{forest}}
    \captionsetup{width=\linewidth}
    \caption{Taxonomy of recent research on precise and efficient test-time compute for large language models.}
    \label{fig:lit_surv}
\end{figure}
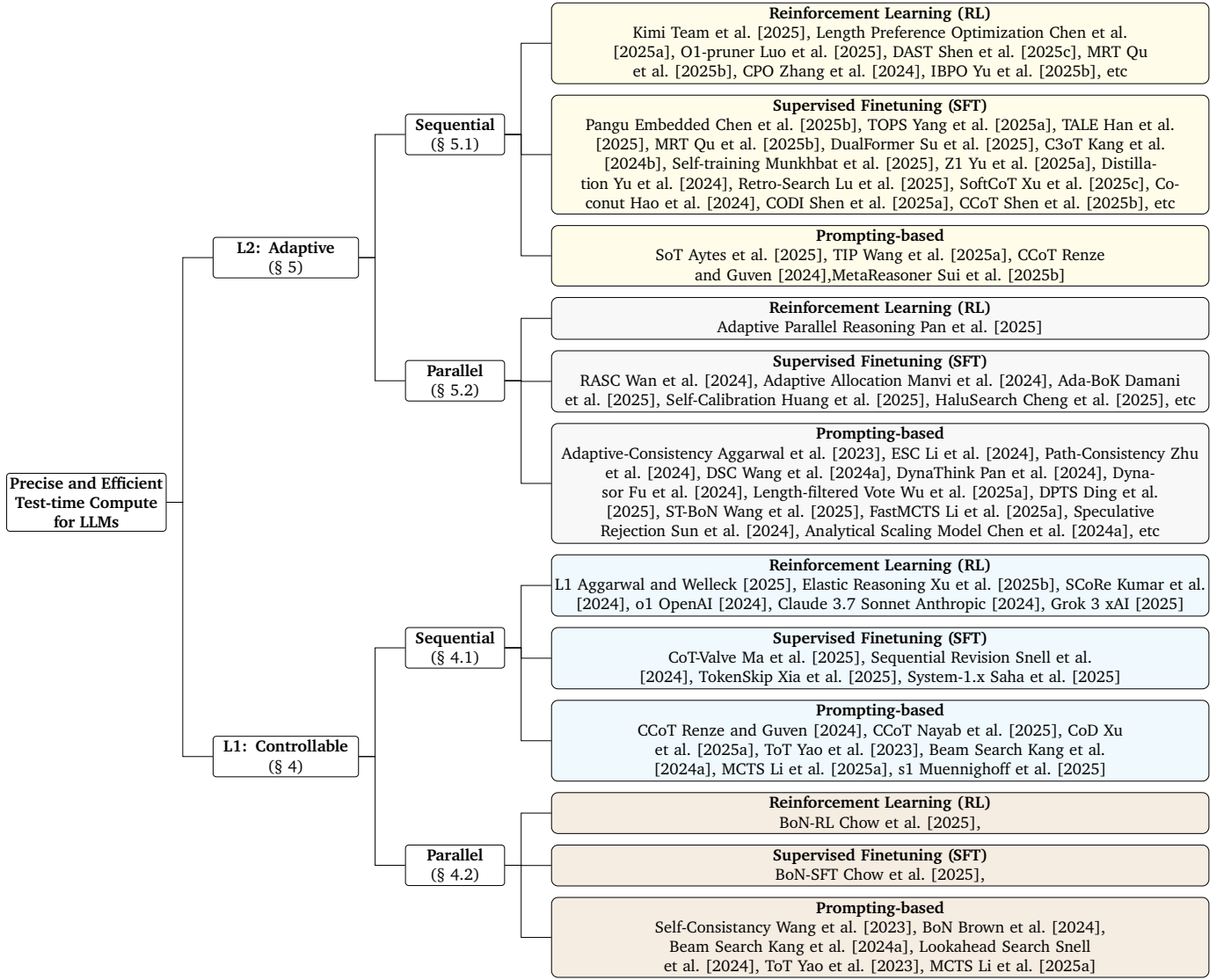

This survey offers a comprehensive review of recent progress in efficient test-time compute (TTC) methods for large language models. In contrast to prior surveys on efficient LLM reasoning~\citep{qu2025surveyefficientreasoning, wang2025harnessingreasoningeconomy, sui2025stopoverthinkingsurvey}, we introduce a new two-tiered framework for characterizing efficiency: controllable (L1) methods, which adhere to fixed inference-time compute budgets; and adaptive (L2) methods, which dynamically adjust computation based on factors such as task complexity and model confidence. We identify and classify the recent progress on inducing efficiency for different test-time compute paradigms (parallel and sequential), using several LLM tuning approaches including prompting, supervised finetuning (SFT), and reinforcement learning (RL). In addition, we benchmark several state-of-the-art proprietary LLMs on diverse reasoning datasets to evaluate their controllability and computational efficiency. By presenting a structured, multi-axis taxonomy (see Figure~\ref{fig:lit_surv}) and in-depth analysis of relevant works, this survey offers both practical guidance and conceptual foundations for future research on efficient test-time compute methods for LLMs. The remainder of this survey is structured as follows. Section~\ref{sec:background} formalizes the notion of test-time compute and introduces our two-tiered taxonomy characterizing controllability (L1) and adaptiveness (L2). Section~\ref{sec:eval-analysis} presents empirical benchmarks on the efficiency of popular TTC methods (i.e., large reasoning models) across multiple datasets. Section~\ref{sec:l1} reviews prior work on controllability (L1), while Section~\ref{sec:l2} surveys methods targeting adaptiveness (L2). Section~\ref{sec:discussion} outlines practical applications and emerging research directions. Section~\ref{sec:conclusion} concludes the survey.

\section{Background}\label{sec:background}

In this section, we provide an overview of the test-time compute paradigm and its variants (Section~\ref{sec:test-time-compute}), introduce the overthinking phenomenon, and formally define reasoning efficiency (Section~\ref{sec:overthinking}).

\subsection{Test-time Compute}\label{sec:test-time-compute}

Test-time compute methods aim to allocate more computation to solving problems at test-time rather than only scaling LLM pre-training~\citep{zhang2025whathowwherewell}. This is akin to the system-2~\citep{kahneman2012thinkingfastslow} thinking that humans employ when solving novel problems. Following prior works~\citep{snell2024scaling,brown2024largelanguagemonkeys,aggarwal2025l1controllinglongreasoning}, we classify test-time compute methods into two main categories: Parallel and Sequential. 

Parallel methods sample multiple solutions from the LLM for a given problem at test time. The final answer is selected by aggregating all solutions. Classic parallel methods include majority voting (self-consistency)~\citep{wang2023selfconsistency}, Best-of-N~\citep{snell2024scaling}, which utilizes a reward model~\citep{lightman2024letsVerifyStep} to select the best answer.  The success of such parallel search methods is correlated with the base model's reasoning capacity (ability to sample the correct solution), the number of solutions sampled, and aggregation effectiveness (ability to select the correct solution). Therefore, performance typically improves as more candidates are sampled before plateauing (at around 128 samples)~\citep{brown2024largelanguagemonkeys, snell2024scaling}. 

Alternatively, test-time compute can be scaled by sequentially refining a single candidate solution. This refinement can occur either holistically—by updating the entire output—or incrementally—by adding additional reasoning steps. A foundational approach in this category is chain-of-thought (CoT) prompting~\citep{wei2022chainofthought}, which encourages LLMs to generate intermediate reasoning steps before producing a final answer. Building on this idea, self-refinement methods~\citep{chen2025iterativedeepeningsampling, snell2024scaling, kumar2024traininglanguagemodels} iteratively improve an initial response by leveraging the model’s self-correction capabilities. More advanced strategies, such as beam search~\citep{snell2024scaling, kang2024mindstarenhancing} and Monte Carlo tree search (MCTS)~\citep{li2025fastmctssimples}, combine elements of sequential and parallel reasoning. Rather than generating a complete solution in a single forward pass, these hybrid methods explore a tree of candidate reasoning paths, selectively expanding the most promising intermediate steps to converge on a high-quality final answer.


Recent large reasoning models (e.g., o1~\citep{openai2024openaio1card} and R1~\citep{deepseekai2025deepseekr1}) are trained with reinforcement learning (RL) to produce long CoTs before arriving at a final answer. These models learn to internalize the search process found in prior sequential and parallel methods (e.g, backtracking, self-evaluation, and parallel exploration) through extensive training. Empirically, such models are found to scale better than pure parallel methods since later computations can build on intermediate reasoning steps, allowing for deeper reasoning and iterative refinement~\citep{aggarwal2025l1controllinglongreasoning, muennighoff2025s1simpletest}. However, harder problems that require non-linear multi-hop reasoning paths (e.g., spatial reasoning~\citep{yamada2024evaluatingSpatialUnderstanding}) may benefit from both scaling paradigms~\citep{snell2024scaling}.

\subsection{Problem Definition}\label{sec:overthinking}

While test-time compute brings immense performance benefits, it comes with a heavy computational cost~\citep{wang2025thoughtsplacethinking, chen2025think23overthinking, aggarwal2025l1controllinglongreasoning}. For example, large reasoning models can produce long CoTs employing extensive and deep thinking, resulting in significant inference overhead, even for simple questions (e.g., 1+1=?)~\citep{chen2025think23overthinking}. Likewise, parallel sampling approaches, despite their improvements with a large number of samples, increase cost proportional to the number of sampled LLM outputs~\citep{han2025tokenbudgetaware}. However, not all problems require such significant test-time computation. This is especially true for problems where LLMs have acquired significant knowledge during pre-training (e.g., one-hop factual questions). Therefore, to obtain optimal efficiency, one would want a system that can automatically switch between fast and slow thinking based on the problem at hand, inspired by the dual process theory~\citep{kahneman2012thinkingfastslow}. The fast thinking relies on intuition and knowledge learned from prior experience. The slow thinking is more costly and relies on deliberate planning (e.g., search and self-reflection), similar to what is employed by o1-like models.

\begin{figure}[h]
    \centering
    \includegraphics[width=\linewidth]{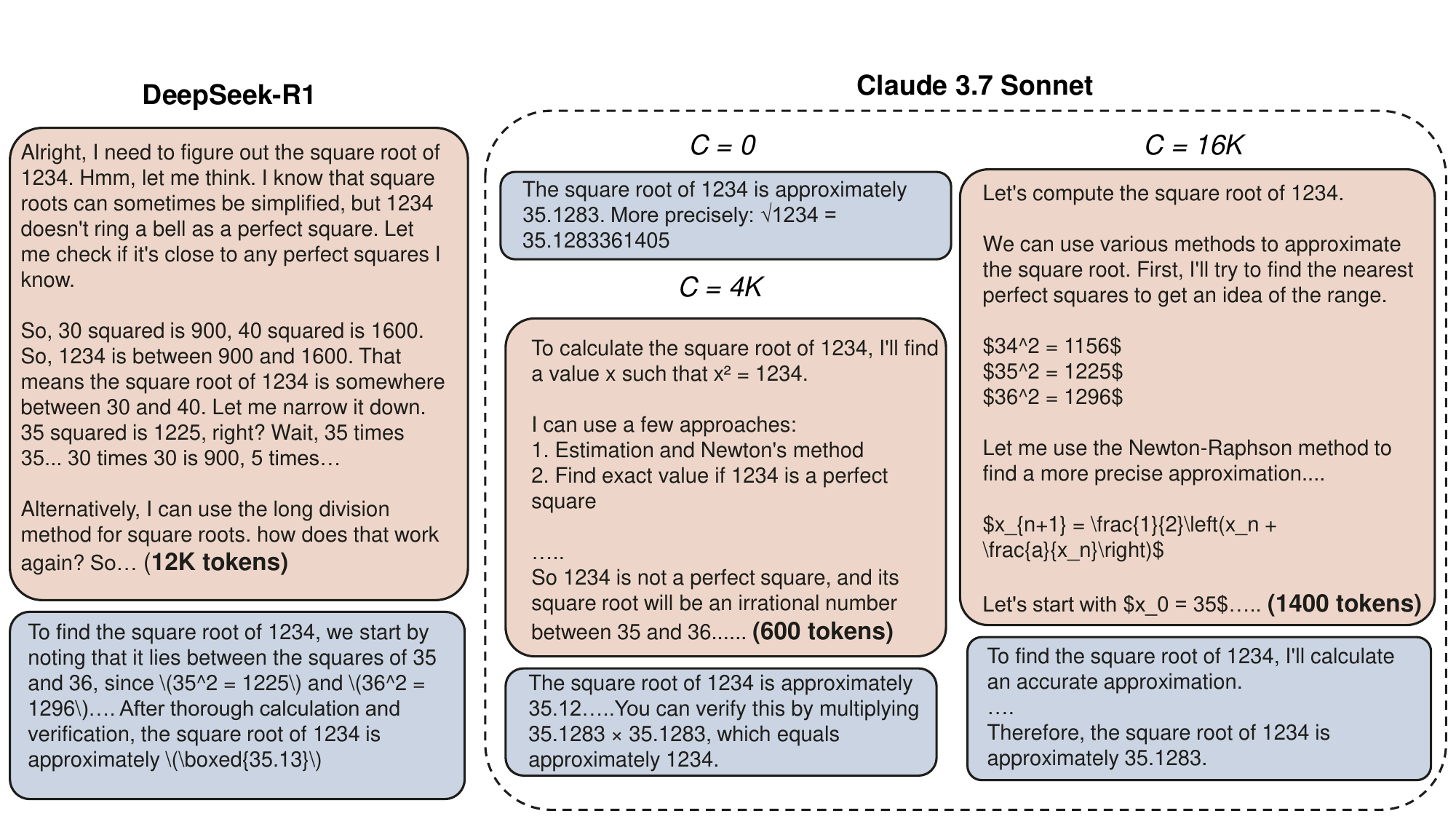}
    \captionsetup{width=\linewidth}
    \caption{Controllable (\textit{Claude 3.7 Sonnet}) vs uncontrollable (\textit{DeepSeek-R1}) large reasoning model outputs for the question: \textit{"What is the square root of 1234?"}. Thinking and answer outputs are represented in \textcolor{brown}{brown} and \textcolor{MidnightBlue}{blue} respectively. $C$ represents the thinking compute budget (i.e. maximum \# of thinking tokens). R1 spends considerable test-time compute (12K tokens) but gives equally reliable answers to the Claude model under lower budget constraints.}
    \label{fig:controllable_vs_uncontrollable}
\end{figure}

Prior works on efficient test-time compute can be classified into two main paradigms: \textit{\textbf{Controllable test-time compute (L1)}} and \textit{\textbf{adaptive test-time compute (L2)}}. Controllable test-time compute methods abide by pre-defined budget constraints set by the user (e.g., number of samples/tokens). See Figure~\ref{fig:controllable_vs_uncontrollable} for an example. In contrast, adaptive test-time compute methods allocate compute budget dynamically, taking into account the problem difficulty and the LLM reasoning capabilities. More formally, let $\mathcal{P}(r, x) \in \mathbb{R}$ and $\mathcal{E}(r, x) \in \mathbb{R}$ be the performance and efficiency metrics, respectively, for a reasoning trace $r \in \mathcal{R}$ generated by a method given a problem $x \in \mathcal{X}$. The reasoning trace represents the full search process of the test-time compute method. For example, for the self-consistency method~\citep{wang2023selfconsistency}, $r$ represents the set of generated responses.
For a pre-defined budget constraint $C \in \mathbb{R}$, L1 methods optimize a constrained optimization problem (equation~\ref{eq:l1_optim}), while L2 methods optimize a penalized function (equation~\ref{eq:l2_optim}) over the set of possible reasoning traces $\mathcal{R}$: 
\begin{align}
&\mathrm{L1: Controllable:} \quad \max_{r \in \mathcal{R}} \,\,\mathcal{P}(r, x)
        \quad \text{ s.t. } \,\,\mathcal{E}(r, x) 
        \geq \frac{1}{C} \label{eq:l1_optim}\\[10pt]       
&\mathrm{L2: Adaptive:} \quad \quad \,\,\max_{r \in \mathcal{R}} \,\, \mathcal{P}(r, x) + \alpha \mathcal{E}(r, x) \label{eq:l2_optim}
\end{align}
where $\alpha \in \mathbb{R}$ is a hyperparameter. Sequential methods typically define the budget as the total number of output tokens, while parallel methods define it as the number of sampled solutions before returning the final response. In this survey, we position the two paradigms within a hierarchy of efficiency, where L2 (adaptive) methods are ultimately more practical than L1 (controllable) methods since they require no pre-defined budget constraints. At inference time, parallel search strategies offer readily controllable leverage over the computational budget and the extent of the search space explored. This control is commonly achieved by adjusting parameters such as the number of parallel samples generated, as seen in methods like self-consistency~\citep{wang2023selfconsistency}, best-of-N~\citep{brown2024largelanguagemonkeys} or MCTS~\citep{li2025fastmctssimples}. 
For iterative prompting-based sequential methods~\citep{zheng2023progressive, palrefining}, we could instead control the number of self-refinement rounds performed, which dictates how deeply the model engages in revisiting and improving its initial answers. 
However, achieving fine-grained control over the dynamic, step-by-step thinking process inherent in sequential generation methods presents a more complex challenge, especially when the model's internal reasoning path has been implicitly linearized or improved via advanced post-training regimes such as reinforcement learning. Moreover, \textit{it is noteworthy that our problem formulation focuses on the efficiency of the reasoning process itself (e.g., reducing number of steps)}. Therefore, inference optimization techniques such as quantization~\citep{frantar2023optqAccurateQuantization}, KV caching~\citep{luohe2024keepCostDown}, and other model compression methods~\citep{wan2024efficientlargelanguage} are out of the scope of this survey as they aim to optimize the inference time, regardless of the LLM response length.

To optimize the reasoning process itself, several metrics have been proposed to quantitatively measure the overthinking phenomenon in test-time compute methods. Token cost measures the total number of tokens generated by the LLM, including all samples and their intermediate reasoning steps~\citep{han2025tokenbudgetaware}. This metric has been widely adopted for both parallel and sequential methods as it is directly correlated to computational and monetary cost. FLOPs-based cost has also been recently adopted to benchmark test-time compute methods under the same compute budget~\citep{snell2024scaling}. Other metrics also take into account the correctness of the responses. For example, the Outcome efficiency metric~\citep{chen2025think23overthinking} $\mathcal{E}_O$ measures the proportion of tokens that contribute to the correctness of an LLM response:
\begin{equation*}
    \mathcal{E}_O(r, x) = \frac{\hat{\mathcal{T}}(r)}{\mathcal{T}(r)}
\end{equation*}
where $\mathcal{T}()$ is the total number of output tokens, and $\hat{\mathcal{T}}()$ is the number of tokens to first arrive at the correct answer in the response $r$. Consequently, any model that discovers the answer at an early stage but continues to generate redundant reasoning steps will be considered inefficient. 

We further categorize efficient test-time compute methods based on how efficiency is achieved. \textit{Prompting-based} approaches adjust the LLM's compute budget without any finetuning. For example, one can instruct the model to generate a response within a specific number of tokens~\citep{han2025tokenbudgetaware, xu2025chainofdraft}. \textit{Supervised finetuning (SFT)} and \textit{reinforcement learning (RL)} approaches typically elicit efficiency by finetuning the LLM on efficient responses or directly optimizing the aforementioned efficiency metrics (e.g. maximizing efficiency metrics as rewards~\citep{team2025kimi}).

\begin{figure}[t]
    \centering
    \includegraphics[width=0.6\linewidth]{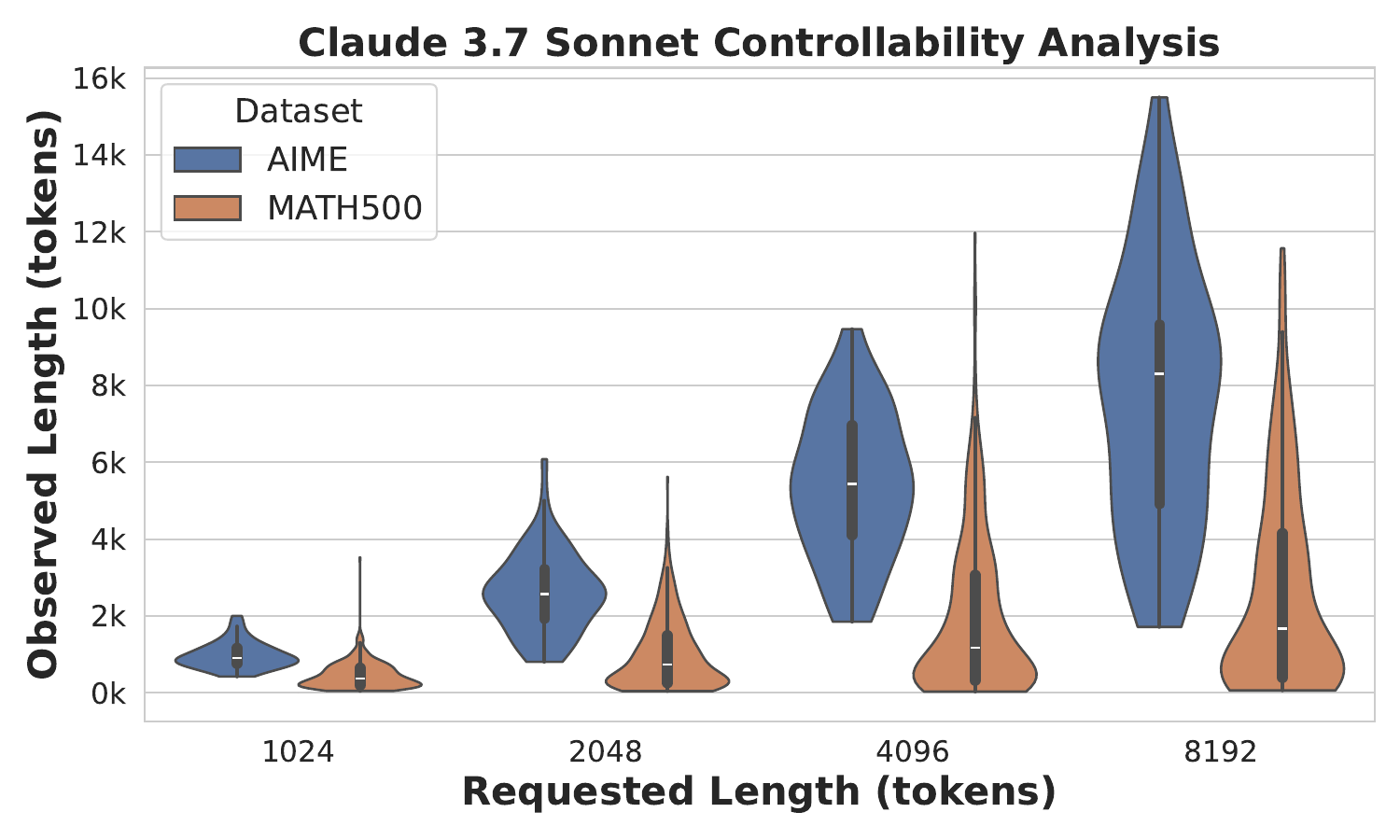}
    \captionsetup{width=\linewidth}
    \caption{Distribution of response lengths for AIME 2024-2025 and MATH500 datasets with Claude 3.7 Sonnet. "Requested Length" represents the thinking token budget set (i.e., number of tokens between the <think></think> tags) for the Claude model. "Observed Length" represents the number of thinking tokens consumed by the model.}
    \label{fig:claude_controllability}
\end{figure}

\begin{figure}[t]
    \centering
    \begin{subfigure}{0.49\textwidth}
        \includegraphics[width=\linewidth]{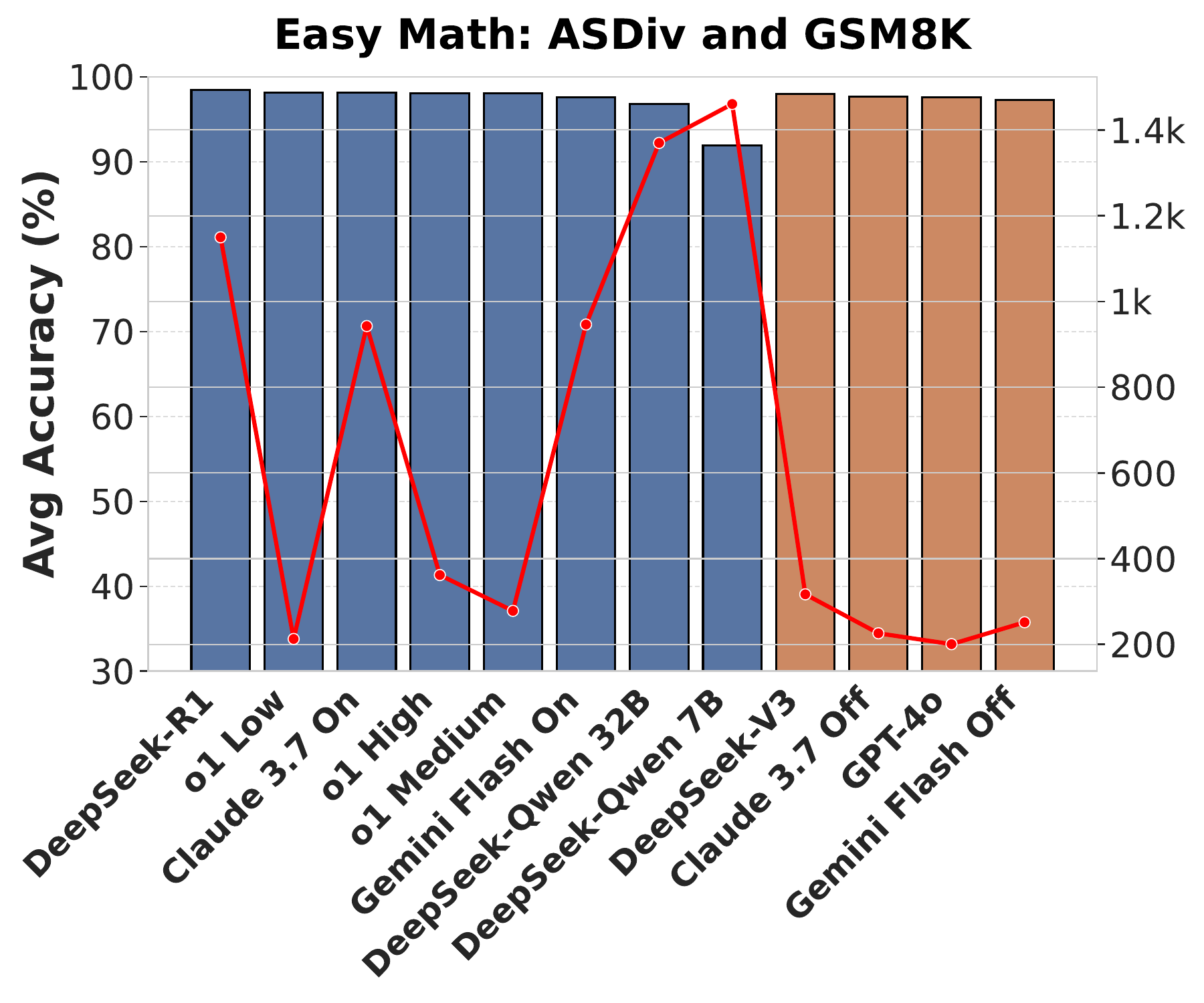}
    \end{subfigure}
     \begin{subfigure}{0.49\textwidth}
        \includegraphics[width=\linewidth]{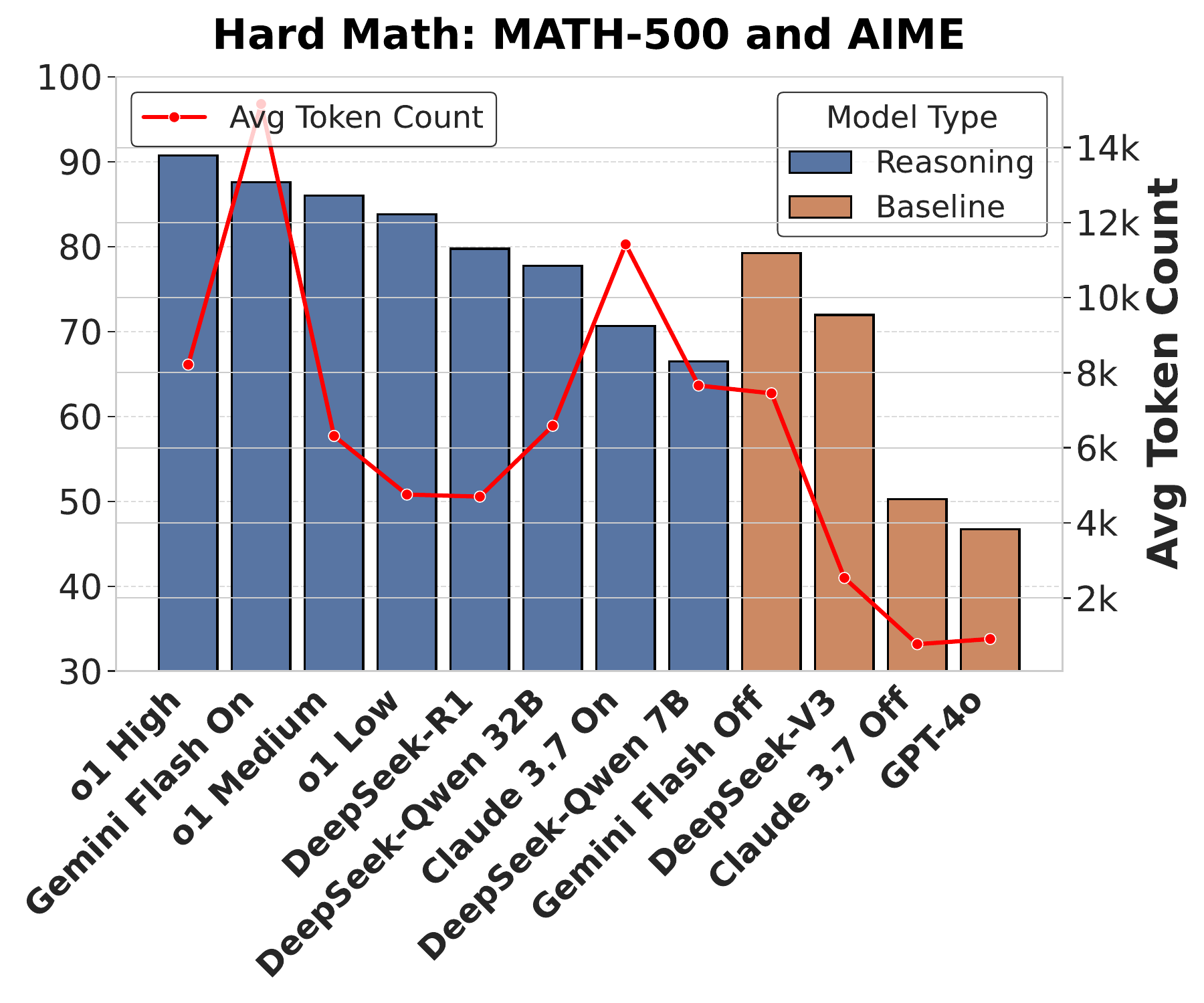}
    \end{subfigure}
     \begin{subfigure}{0.49\textwidth}
        \includegraphics[width=\linewidth]{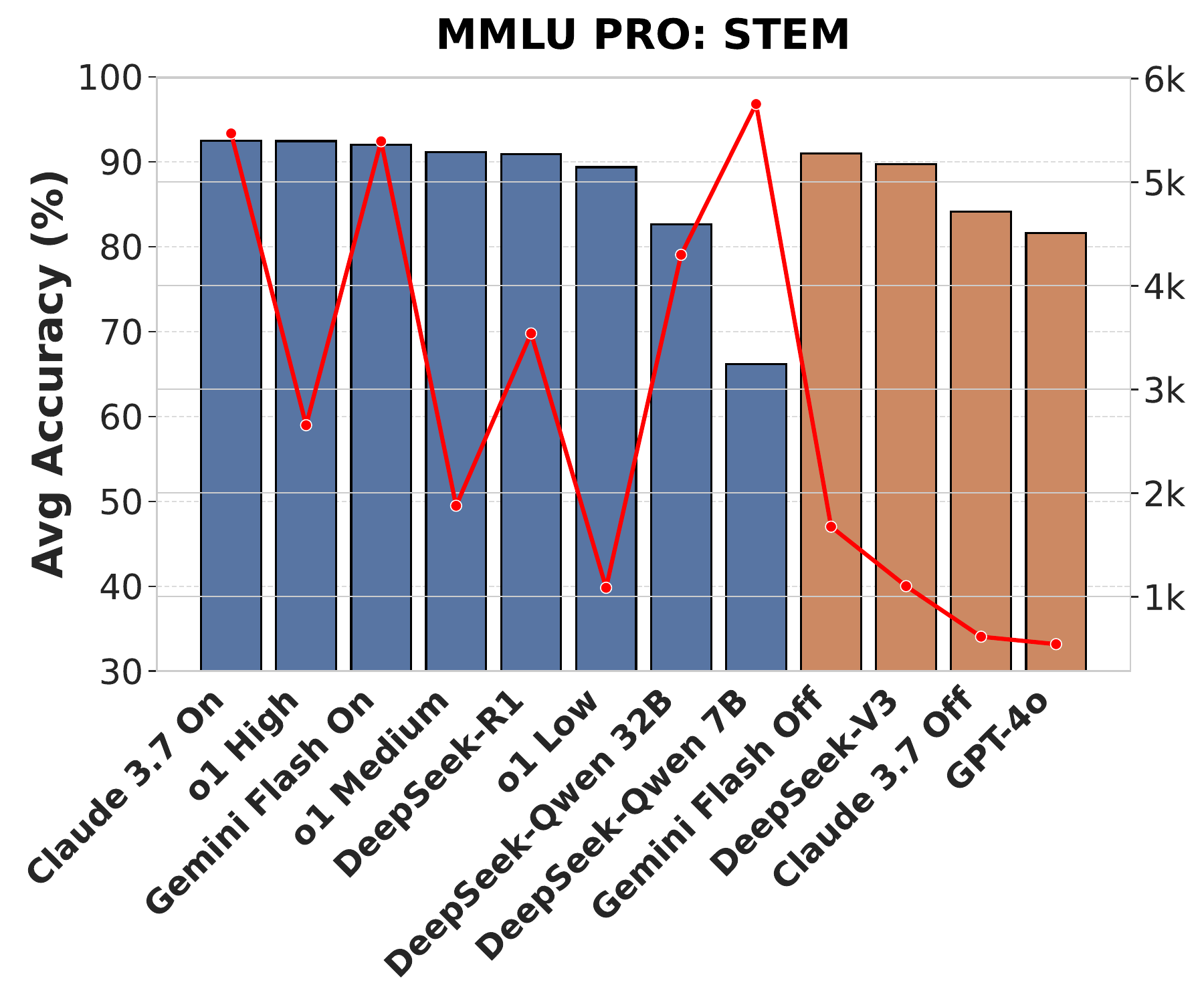}
    \end{subfigure}
     \begin{subfigure}{0.49\textwidth}
        \includegraphics[width=\linewidth]{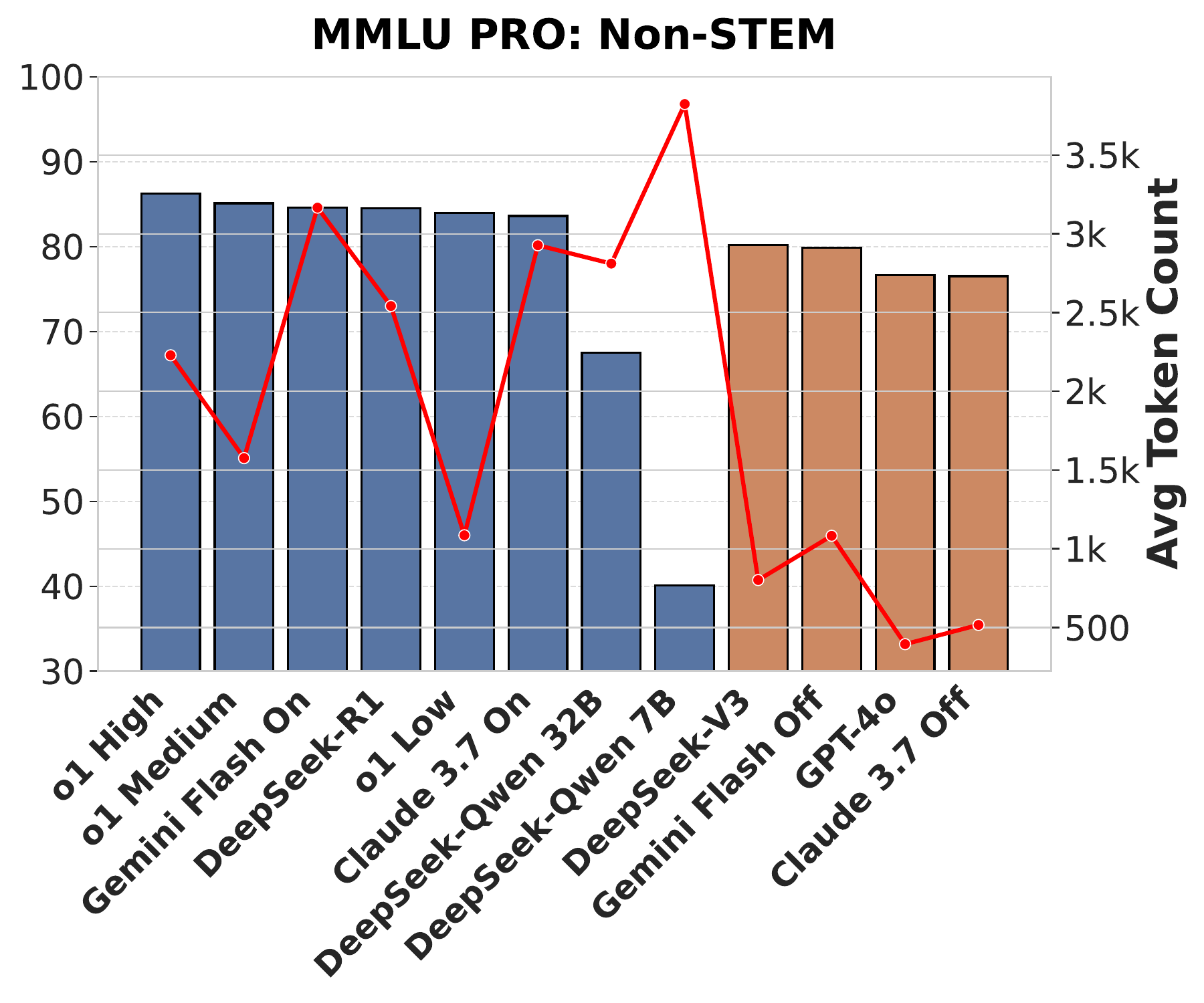}
    \end{subfigure}
    \captionsetup{width=\linewidth}
    \caption{Average Pass@1 accuracy and average Token Count for reasoning and non-reasoning (baseline) models on different dataset categories. MMLU Pro STEM contains Physics, Biology and Chemistry questions. MMLU Pro Non-STEM contains History, Law, and Business questions. \texttt{Gemeni Flash Off} and \texttt{Claude 3.7 Off} represent LLMs with ``thinking'' mode turned off.}
    \label{fig:model_perf_acc_token_count}
\end{figure}

\section{Efficiency of Large Reasoning Models}\label{sec:eval-analysis}


To ground our taxonomy and highlight the practical limitations of current LLMs, we empirically evaluate the efficiency and controllability of leading open-source and proprietary reasoning models. While recent advances in test-time compute (TTC) have enabled substantial gains in reasoning accuracy, they often come at the cost of excessive and poorly calibrated inference-time computation. In this section, we assess how well existing models align with our two-tiered framework—L1 controllability and L2 adaptiveness—by analyzing their ability to modulate computation with respect to user-specified budgets and task difficulty.

Section~\ref{sec:controllability_exp} investigates L1 controllability in commercial models such as Claude 3.7, focusing on how precisely the model adheres to an explicit ``thinking budget'' across datasets of varying difficulty. Section~\ref{sec:effiency_exp} complements this with a broad benchmark of LLMs in terms of token usage and reasoning performance across four dataset categories. These experiments reveal systemic inefficiencies in current TTC strategies, including underthinking on hard problems, overthinking on simple ones, and limited sensitivity to task complexity—all of which underscore the need for more adaptive and compute-aware reasoning mechanisms. More details on the experimental setup can be found in Appendix~\ref{app:exp_setup}.


\subsection{Controllability Analysis for Existing Models}\label{sec:controllability_exp}

In line with our L1 problem formulation, recent large reasoning models~\citep{anthropic2024claude37systemcard,openai2024openaio1card} have the ability to control the amount of thinking (computation) allocated at inference time. For example, the o-series~\citep{openai2024openaio1card,openai2025o3} support a ``reasoning effort'' parameter that can be set to low, medium, or high to control the amount computation a user wants the model to spend. Anthropic's Claude 3.7 Sonnet~\citep{anthropic2024claude37systemcard} supports a ``thinking budget'' parameter that sets the exact amount of thinking tokens to be spent on the given prompt. 

In Figure~\ref{fig:claude_controllability}, we assess Claude's ability to abide by the thinking budget on the AIME~\citep{aime2025artofproblemsolving} and MATH500~\citep{hendrycks2021measuringMathematicalProblem} datasets. Unsurprisingly, the results show that the model spends significantly more tokens for AIME problems due to their difficulty. However, while the observed thinking length grows proportionally to the allocated budget, the results show a long tail of problems where the model significantly exceeds the budget. This suggests that more work is needed to enable consistent budget control for such large reasoning models. We discuss a few recent works that tackle this problem in Section~\ref{sec:l1_sequential}.





\subsection{Efficiency Analysis for Existing Models}\label{sec:effiency_exp}



Figure~\ref{fig:model_perf_acc_token_count} benchmarks the average performance and response lengths (in tokens) of a range of LLMs across four dataset categories. The results reveal a consistent trend: non-reasoning baseline models exhibit significantly lower token usage, particularly on math and STEM tasks. In contrast, high-performing reasoning models trained with reinforcement learning—such as \texttt{o1}\citep{openai2024openaio1card} and \texttt{DeepSeek-R1}\citep{deepseekai2025deepseekr1}—generate responses that are up to \textit{5$\times$ longer}, but achieve \textit{state-of-the-art accuracy} on challenging math benchmarks. Compact reasoning models trained via distillation (e.g., \texttt{DeepSeek-Qwen 7B}) tend to produce the \textit{longest outputs} in three out of four categories, yet underperform relative to baseline models. This aligns with prior findings\citep{chu2025sftmemorizesRLgeneralizes}, which argue that distillation through supervised finetuning can lead to rote memorization of reasoning traces, impairing generalization to out-of-distribution problems.

These results further highlight key challenges in managing test-time compute: (1) sequential reasoning models often lack \textit{budget awareness}, leading to inefficient token usage. (2) On simpler instances, increasing the thinking budget yields diminishing returns and may induce \textit{overthinking} (also observed in Figure~\ref{fig:claude_controllability} for MATH500), whereas harder problems demand a higher compute budget—\textit{underthinking} is frequently the bottleneck. This suggests the need for models capable of \textit{adaptive budget allocation}, modulating computation based on input difficulty and model capabilities. (3) Notably, this overthinking phenomenon is not limited to STEM; it generalizes to broader reasoning tasks (e.g., MMLU Pro Non-STEM), underscoring the importance of controllable and efficient reasoning mechanisms.

\section{Controllable Test-time Compute (L1)}\label{sec:l1}

In this section, we survey recent efforts towards developing controllable test-time compute methods for LLMs. These methods allocate a pre-defined compute budget to solve problems at inference-time. For the sequential methods covered in Section~\ref{sec:l1_sequential}, inference compute budget is measured by the number of tokens or steps in a single response. On the other hand, parallel methods in Section~\ref{sec:l1_parallel} represent the compute budget as the number of sampled responses.

\subsection{Sequential}\label{sec:l1_sequential}

\paragraph{Prompting:} Chain-of-thought (CoT) prompting~\citep{wei2022chainofthought} has emerged as a popular prompting-based test-time compute method that encourages LLMs to arrive at an answer through intermediate reasoning steps. Despite its benefits, CoT prompting can lead to unnecessarily verbose outputs and higher latency, even for simple questions~\citep{renze2024BenefitsConciseChain}. Consequently, several works have proposed prompting-based techniques to control the amount test-time compute that sequential methods spend. \citet{nayab2025concisethoughtsimpact} instruct the LLM to remain below a given word budget (e.g., ``limit the answer to $k$ words''). The authors also propose several efficiency and conciseness metrics and test their method with different word budgets across several models. The results indicate that the small models fail to abide by the budget but larger models can generate shorter responses and sometimes achieve better performance. Alternatively, \citet{xu2025chainofdraft} find that instructing the model to remain below a certain word budget \textit{pre-step} (e.g., ``limit each thinking step to 5 words at most'') allows for unlimited reasoning steps. This reduces token usage (up to 92.4\%) while maintaining accuracy across arithmetic, commonsense, and symbolic reasoning tasks.
Moreover, enforcing a per-step constraint makes it more adaptable to various structured reasoning techniques (e.g., Tree-of thoughts~\citep{yao2023treeofthought}). While the results show positive efficiency improvements for simple arithmetic and reasoning datasets (e.g., GSMK), the results also indicate that the per-step budget constraint brings significant degradations for smaller and weaker models.  \citet{muennighoff2025s1simpletest} introduce a method called ``budget forcing'' to control the minimum and maximum amount of thinking tokens a language model uses during testing. Contrary to \citet{xu2025chainofdraft}'s approach, the authors do not instruct the LLM to constrain its response length. Rather, the method either truncates the model's reasoning process at a predefined maximum point, or prolongs it by appending the prompt with ``Wait'' to encourage the model to reevaluate or refine its reasoning before concluding. The results show better controllability for hard math reasoning datasets. However, the budget forcing method requires careful prompt engineering for the choice of self-reflection tokens (e.g., ``Wait'') to encourage the model to refine its answer

\paragraph{Supervised Finetuning:} To attain better controllability of LLM reasoning, \citet{ma2025cotvalvelengthcompressible} introduce a LoRA adapter~\citep{hu2022loraLowRank} module that controls the length of the CoT. By tuning the weight of this adapter module, the user can control the length of the CoT (e.g. going from system 1 to system 2). The authors construct a dataset of varying CoT lengths to finetune this adapter module. Experiments show greater controllability and efficiency over other RL-based approaches that encourage efficiency. \citet{xia2025tokenskipcontrollable} analyze the semantic importance of tokens in CoTs and find that most tokens are redundant to reasoning. Building on this insight, the authors introduce TokenSkip, a controllable CoT compression method for LLMs. TokenSkip first generates CoT trajectories from the target LLM. These CoTs are then compressed to a specified ratio, based on the semantic importance of tokens. TokenSkip fine-tunes the LLM using the compressed CoTs, allowing for efficient inference with adjustable compression ratios. Unlike \citet{ma2025cotvalvelengthcompressible}'s compression, TokenSkip generates compressed responses that are difficult to comprehend. The authors demonstrate that the original CoT can be recovered via LLM prompting but this incurs additional inference overhead.

Other works introduce controllability and efficiency for search and planning problems. \citet{saha2025system1xlearning} proposes the System-1.x Planner, a hybrid LLM-based framework that combines fast planning (System-1) and slow, deliberate planning (System-2) to optimize long-horizon
planning tasks. System-1.x is composed of three modules, all of which are implemented with LLMs: controller, system-1 planner, and system-2 planner. The controller decomposes the given problem into sub-goals, classifying them as easy or hard based on complexity. Easy sub-goals are handled by the system-1 planner for efficiency, while hard ones are solved by the system-2 planner using detailed search-based strategies. A user-defined parameter called the hybridization factor allows for controllable adjustment between the two modes, balancing accuracy and computational cost. The three modules are trained via supervised finetuning on ground truth trajectories, generated by symbolic planners (e.g., $A^{*}$).

\paragraph{Reinforcement Learning (RL):} \citet{aggarwal2025l1controllinglongreasoning} propose one of the few methods that permits precise control of LLM reasoning length via reinforcement learning. The method, called ``Length Controlled Policy Optimization (LCPO)'', trains models to adhere to user-specified length constraints. A reward penalty is introduced to force the model to abide by the token budget. Using LCPO, the authors developed two models—L1-Exact, which generates reasoning outputs that match target lengths exactly, and L1-Max, which ensures outputs remain within a maximum length constraint. Experiments show good adherence to token budget after RL training and a linear increase in performance as token budget is increased. However, it is unclear if L1 can generalize to larger context sizes as the experiments are limited to a maximum thinking length of 3600 tokens. 

Rather than including length instructions in the prompt, \citet{xu2025scalablechainthoughtselastic} truncate the thinking length to meet a given budget and train the LLM under these constraints via reinforcement learning. Moreover, unlike L1~\citep{aggarwal2025l1controllinglongreasoning} which only considers the thinking budget, the method divides the total token
budget $c = t + s$ into two parts: $t$ tokens for the thinking phase and s tokens for the solution phase. The LLM thinking phase is terminated (i.e, by appending ``</think>'' tag) once $t$ tokens are consumed. The authors also introduce \textit{budget-constrained rollout}, which teaches the model to generate correct answers even with partial thinking trajectories. 

\subsection{Parallel}\label{sec:l1_parallel}

\paragraph{Prompting:} 

Querying the LLM multiple times in parallel to generate multiple solutions to the same problem and aggregating those answers is another way to improve reasoning. To that end, \citet{wang2023selfconsistency} propose sampling of a fixed number of CoTs for each task and conduct a majority vote among extracted answers for solving a reasoning task. The resulting method is termed \emph{self-consistency}.
While this approach can be applied to tasks with only one correct answer, significant modifications are necessary for applying it to tasks with free-form answers. \citet{chen2024universal} propose \emph{universal self-consistency}, which uses a subsequent LLM call to obtain the majority consensus for the final answer in such tasks (e.g., code generation, abstractive
summarization, and open-end question answering). A similar approach is proposed in~\citet{wang2024integrate}.  To alleviate cases where the majority is wrong, Best-of-N (BoN) methods~\citep{cobbe2021trainingverifierssolve,brown2024largelanguagemonkeys,lightman2024letsVerifyStep,liu2025can1bllmsurpass405b} leverage a verifier or reward model to score the candidate samples before aggregating the scores (e.g., max) to pick the best candidate.

Parallel methods leverage the intuition that problems often have multiple possible solutions with difference reasoning paths that reach the same unique answer. Unlike sequential methods, compute budget $C$ is typically defined as the number of samples. It has been found that such repeated sampling methods can balance cost and performance as it can amplify a weaker model’s capabilities without finetuning and outperform single attempts from bigger models~\citep{liu2025can1bllmsurpass405b, brown2024largelanguagemonkeys}.

\paragraph{Finetuning (SFT and RL):} Standard LLMs are trained to produce the best response for a given prompt. That is, LLMs are trained to maximize the likelihood of a ground truth response (supervised finetuning) or maximize an environment reward (reinforcement learning). While successful, these training strategies do not take the aforementioned test-time compute methods (e.g., self-consistency~\citep{wang2023selfconsistency}) into account. \citet{chow2025inferenceawarefinetuning} present an \textit{inference-aware} finetuning paradigm that explicitly considers the inference procedure during training. In particular, the authors study this inference-aware paradigm under the Best-of-N (BoN) strategy. Several SFT (BoN-SFT) and RL (BoN-RL) methods are presented for BoN-aware LLM training. Empirical results demonstrate that this inference-aware training improves the exploitation-exploration tradeoff (i.e., interleaving diverse and high-likelihood responses); therefore, improving efficiency. However, it is unclear how their method can be applied to TTC methods beyond BoN such as self-consistency~\citep{wang2023selfconsistency} and MCTS~\citep{li2025fastmctssimples}.

\begin{tcolorbox}[
  enhanced,
  colback=white,
  colframe=black,
  boxrule=1.0pt,
  fonttitle=\bfseries,
  title=Closing Insight on L1 Methods,
  attach boxed title to top left={yshift=-2mm, xshift=2mm},
  boxed title style={
    colback=MidnightBlue,
    colframe=black,
    boxrule=0pt,
    fontupper=\color{white}\bfseries
  }
]
Controllable test-time compute offers a promising avenue for tailoring LLM inference to specific efficiency needs, enabling users to trade off performance and cost in a principled way. Sequential methods offer better controllability in terms of the number of tokens However, they often require task-specific tuning or manual prompt engineering, limiting their generality. \textit{A key challenge moving forward is to develop more robust and intuitive token-level control mechanisms that generalize across models, tasks, and compute budgets.}
\end{tcolorbox}

\section{Adaptive or Near-optimal Test-time Compute (L2)}\label{sec:l2}

In this section, we focus on test-time compute methods that can adaptively allocate compute at inference-time. Contrary to the L1 methods explored in Section~\ref{sec:l1}, L2 methods do not require pre-defining a compute budget by the user. As in Section~\ref{sec:l1}, we classify L2 methods into sequential (Section~\ref{sec:l2_sequential}) and parallel (Section~\ref{sec:l2_parallel}) methods.

\subsection{Sequential}\label{sec:l2_sequential}

\paragraph{Prompting:} Several works have explored eliciting adaptive efficiency through simple LLM prompts. \citet{renze2024BenefitsConciseChain} introduce Concise Chain-of-Thought (CCoT) prompting, which is achieved by simply instructing the LLM to both ``think step-by-step'' and ``be concise''. The study compares standard CoT and CCoT prompting using GPT-3.5 and GPT-4 on multiple-choice benchmarks. The results show that CCoT reduces response length without significantly impacting accuracy, although a performance penalty is observed in math problems for weaker models (e.g., GPT-3.5).  \citet{aytes2025sketchofthought} propose three broad reasoning approaches that encourage efficiency inspired by cognitive science: constructing reasoning steps with minimal conceptual chains (Conceptual Chaining), using symbols
or compact representations instead of verbose language (Chunked Symbolism), and employing domain-specific terminology to enhance efficiency (Expert Lexicons).  Additionally, the paper introduces a lightweight routing model that dynamically selects the optimal reasoning approach based on the input problem type. Empirical results show that this method reduces output token usage to just 15-30\% of the original while maintaining high accuracy. However, it is unclear if the proposed cognitive reasoning procedures can generalize beyond simple math and logical reasoning datasets. 

\citet{lee2025llmscompresschainofthought} benchmark a myriad of prompting techniques that elicit efficiency, introduced by prior works~\citep{renze2024BenefitsConciseChain, jin2024impactreasoningstep, nayab2025concisethoughtsimpact}, such as ``be concise'', ``Only use bullet points'', and ``respond in Chinese''. The authors introduce the concept of ``token complexity'', which refers to the minimal number of tokens required to solve a problem effectively. They highlight a tradeoff between compression and performance: while shorter reasoning chains reduce token use, they also often reduce accuracy. The paper emphasizes that current compression methods do not operate near the theoretical limits of efficiency, suggesting room for improvement. Key insights in this paper include the importance of adaptive efficiency, where simpler tasks receive more compressed responses. Moreover, it is found that LLMs can natively adjust response length to problem difficulty, even without explicitly being prompted to do so. Surprisingly, the authors find that this capability does not need sophisticated prompting strategies. Nonetheless, the paper does not benchmark methods based on fine-tuning~\citep{kang2024c3otgeneratingshorterchain, han2025tokenbudgetaware}  which may be able to outperform these simple prompting strategies (See SFT paragraph). \citet{sui2025metareasonerdynamic} introduce a more sophisticated prompting framework known as MetaReasoner. MetaReasoner improves reasoning efficiency and accuracy in LLMs by acting as a strategic advisor during inference. Inspired by human meta-cognition, Meta-Reasoner evaluates reasoning progress iteratively through concise ``progress reports'' generated by the LLM. Using a contextual multi-armed bandit algorithm, it selects optimal prompting strategies such as restarting, backtracking, or simplifying tasks to dynamically guide the reasoning process. By focusing on the high-level strategies rather than step-by-step micromanagement of the reasoning process, Meta-Reasoner mitigates error propagation and reduces wasted computation.

The aforementioned works typically measure and optimize LLM reasoning efficiency on the \textit{correct responses}. Instead, \citet{wang2025thoughtsplacethinking} propose to evaluate the token efficiency within \textit{incorrect responses}. Through extensive analyses of incorrect responses, the authors discover a peculiar phenomenon known as underthinking, where o1-like LLMs tend to prematurely abandon promising lines of reasoning, leading to frequent switching between different solution strategies without reaching a conclusion. To mitigate this problem, the paper proposes decoding and prompting-based methods that explicitly discourage or penalize ``thought'' tokens (e.g., ``Alternatively''). Experiments show this effectively reduces underthinking and enhances accuracy across difficult mathematical and scientific problems.

\begin{tcolorbox}[
  enhanced,
  colback=white,
  colframe=black,
  boxrule=1.0pt,
  fonttitle=\bfseries,
  title=Prompting Can Modulate Reasoning Depth Without Fine-tuning,
  attach boxed title to top left={yshift=-2mm, xshift=2mm},
  boxed title style={
    colback=MidnightBlue,
    colframe=black,
    boxrule=0pt,
    fontupper=\color{white}\bfseries
  }
]
LLMs can dynamically regulate their reasoning depth and output length in response to task complexity—often without any model updates. Simple directives like “be concise” or “use bullet points” can meaningfully compress reasoning chains, especially on easier tasks. However, these strategies often face diminishing returns on harder problems, particularly with weaker models. Notably, \textit{prompting alone may not prevent pathological behaviors like underthinking or premature strategy switching} (See Figure~\ref{fig:claude_controllability} for an example). These findings highlight both the promise and limits of prompt-based efficiency, and underscore the need for more structured control over reasoning behavior.
\end{tcolorbox}

\paragraph{Supervised Finetuning (SFT):} While prompting-based techniques can elicit efficiency reasoning, their evaluation has been limited to few models. \citet{munkhbat2025selftrainingelicits} find that such methods cause accuracy reduction on small task-specific models. To tackle this issue, the paper proposes to finetune LLMs on concise reasoning data generated from two sources: Best of $N$ and few-shot conditioning. Best of $N$ takes $N$ samples and selects the shortest reasoning trace with the correct answer. Few-shot conditioning leverages human annotated few-shot prompts to  demonstrate very concise reasoning to the model. Experiments show that self-training on these outputs internalizes efficient reasoning, allowing the model to reason concisely without additional inference-time overhead. \citet{yang2025thinkingoptimalscaling} instructs the LLM to generate CoTs of different reasoning efforts (low, medium, high) for each question. The LLM is finetuned on these responses in order to generate CoTs of controllable length. An iterative self-improvemant SFT scheme is also proposed so that the model outputs the shortest CoT possible for a given question. Other works \citep{kang2024c3otgeneratingshorterchain,chen2025panguembeddedefficient} take a similar approach but finetune the LLM on CoTs of two reasoning efforts only (short and long). The short CoTs are generated by prompting the LLM to compress original long CoTs into corresponding short CoTs and adding special tokens before each CoT to indicate its type. A mixed SFT dataset of long and short CoTs is then constructed to help the model learn both reasoning styles and their relation. During inference, inserting the special token for short CoTs guides the model to generate a shorter yet equally effective reasoning process. The LLM can also adaptively decide the right reasoning effort for the given prompt.

\citet{han2025tokenbudgetaware} propose to dynamically allocate the token budget for O1-like models. The authors propose two variants of this approach TALE-EP (estimation+prompting) and TALE-PT (post-training). Given a question, TALE-EP uses the reasoning LLM itself with a zero-shot estimation prompt as the token budget estimator. Using this estimate, it then crafts a token-budget-aware prompt and feeds it into the LLM to generate the final answer. TALE-PT trains the LLM to produce efficient answers without budget estimation. This is done by finetuning the LLM on question-answer pairs with the optimal token budget. Rather than allocating token budget, \citet{liu2024canLanguageModels} teach LLMs to dynamically allocate the number of reasoning steps in a CoT. The training pipeline consists of two stages: first, the model is initialized with a full CoT and a manually constructed CoT with fixed step skipping; then, iterative training is performed where the model samples shorter but correct reasoning paths, which are used to continuously fine-tune the model, enabling it to gradually learn how to dynamically decide which steps to skip based on the input problem. \citet{su2025dualformerControllableFast} train a transformer model adaptively skip steps when solving path-finding problems. The model dynamically switches between fast (System 1) and slow (System 2) reasoning mode. The framework achieves this by randomly dropping parts of the reasoning trace during training, allowing the model to learn both reasoning strategies and autonomously determine when to engage in fast or slow reasoning during inference.

Some works leverage distillation to improve efficiency. \citet{yu2024distilling21} explore the simple idea of distilling the outputs of a System 2 method~\citep{saha024branchsolvemerge, wei2022chainofthought} into a System 1 method (e.g. input/output prompting), thereby transferring the knowledge and reasoning capabilities to the more instinctive and efficient models. Although this distillation can be done directly, they found it critically important to additionally filter System 2 outputs by quality (i.e., self-consistency) before distilling them into a System 1 model. Experiments show significant performance improvements for System 1 methods while achieving better token efficiency. \citet{lu2025retrosearchexploring} propose to revise the reasoning paths outputted by large reasoning models~\citep{deepseekai2025deepseekr1, openai2024openaio1card} before performing distillation to smaller models. Their method, Retro Search, revises the modal outputs to discover better, yet shorter traces. This is done by performing MCTS-like search over alternative solution paths to discover more efficient solutions. Finetuning on these solutions can then lead to student models with enhanced reasoning capabilities with shorter and faster inference. Rather than dynamically searching for shorter reasoning trajectories, \citet{zhao2025letllmsbreakfree} truncate inefficient solutions generated by teacher model at ideal breaking points. The breaking points are identified via a set of ovethinking metrics. Moreover, the authors append breaking prompts (e.g., ``Wait, I've verified my answer. No need to continue thinking'') at the identified braking points, giving the student model the ability to recognize when to end a reasoning process. \citet{yu2025longshortchainofthought} introduce another CoT rewriting method where a LLM is used to compress the long CoT responses into shorter ones without modifying the structure. The authors found that training the student model on both long and short CoT data gives the best tradeoff between efficiency and accuracy.

To enforce a ``think-before-answer'' pattern, existing reasoning models make use of delimiters (e.g., ``<think> ... </think>'') to split the response into the reasoning part and final answer.  \citet{yu2025z1efficienttesttimesscaling} find that this approach causes unnecessary reasoning when processing simple problems that do not require deep thought. Therefore, the authors introduce the Shifted Thinking Window method, where thinking delimiters are eliminated and the maximum answer length is capped. If the model exceeds the cap, the response is interrupted and appended with a hint phrase to enforce a direct answer.

Another line of research explores avoiding decoding to discrete tokens at intermediate steps~\citep{cheng2024compressedchainthought, hao2024traininglargelanguage}. Such methods conduct reasoning in a continuous space by using latent representations instead of discrete token sequences, thereby alleviating the need for unnecessary language tokens used for fluency. \citet{hao2024traininglargelanguage} propose to feed the last hidden layer representations of the LLM as input embedding for the next token rather than mapping it back to discrete token space. During training,
the discrete CoT tokens are gradually replaced with continuous thoughts while maintaining the language modeling objective on the remaining CoT tokens.
This curriculum learning strategy encourages the continuous representations to behave like the discrete CoT tokens, but frees the reasoning from being within the language space, thereby, generating significantly fewer tokens during inference. However, \citet{xu2025softcotsoftchain} show that this method suffers from catastrophic forgetting and requires full model finetuning. Therefore, the authors introduce a novel approach where a weaker assistant language model generates latent-space thoughts, which are then projected into the main LLM for inference. While the projection module requires training, both the assistant and main LLM remain frozen, preserving generalization and avoiding catastrophic forgetting. \citet{shen2025codicompressingchain} compress CoT reasoning into a continuous latent space using self-distillation. A single shared model functions as both the teacher and the student: the teacher learns from the ground truth discrete token CoT to generate the final answer, while the student generates continuous thoughts before predicting the final answer. \citet{shen2025efficientreasoninghidden} proposes the Heima framework, which improves reasoning efficiency in multi-modal large language models by compressing Chain-of-Thought (CoT) reasoning into compact “thinking tokens” in the hidden space. The Heima Encoder encodes the input problem and generates a continuous reasoning representation, while the Heima Decoder can reconstruct the full textual reasoning when needed, balancing efficiency and interpretability. The authors adopt a similar curriculum learning strategy to the one used in \citep{hao2024traininglargelanguage}, where discrete CoT tokens are progressively replaced with continuous tokens during training.

\begin{tcolorbox}[
  enhanced,
  colback=white,
  colframe=black,
  boxrule=1.0pt,
  fonttitle=\bfseries,
  title=Continuous Representations as a Path to Efficiency,
  attach boxed title to top left={yshift=-2mm, xshift=2mm},
  boxed title style={
    colback=MidnightBlue,
    colframe=black,
    boxrule=0pt,
    fontupper=\color{white}\bfseries
  }
]
Shifting reasoning from discrete tokens to continuous latent spaces offers a promising frontier for test-time efficiency. \textit{These approaches reduce unnecessary decoding and improve throughput without heavily compromising reasoning quality}. However, continuous methods often suffer from catastrophic forgetting, limited interpretability, and dependence on full-model finetuning. Ultimately, the key challenge is achieving efficiency gains without sacrificing traceability or introducing brittleness in reasoning behavior.
\end{tcolorbox}

\citet{zhang2025lightthinkerthinking} propose LightThinker, a method that trains large language models to dynamically compress intermediate thoughts into concise internal hidden states during reasoning, enabling subsequent generation based on the compressed content. The approach reconstructs training data with special tokens and designs custom attention masks to guide the
model on when and how to compress, and how to reason based on compressed information. Since the model only attends to the compressed tokens, this reduces peak memory usage and inference time, while maintaining competitive accuracy. \citet{yan2025inftythinkbreaking} extend this online compression idea but operates at the language level. The authors transform monolithic reasoning CoTs into an iterative process with intermediate summarization. By replacing short reasoning segments with concise progress summaries, they enables unbounded reasoning depth while maintaining bounded computational costs. This approach creates a characteristic saw-tooth memory pattern, significantly reducing computational complexity compared to traditional methods. Meanwhile, \citet{chen2025innerthinkingtransformer} introduce the Inner Thinking Transformer (ITT), designed to efficiently enhance the reasoning capabilities of LLMs without increasing their parameter count.  ITT includes 3 key components: 1) Residual Thinking Connection (RTC), allows the model to iteratively refine token representations by accumulating outputs from multiple processing steps within a layer; 2) Adaptive Token Routing (ATR), employs a routing network to dynamically allocate computational resources to tokens based on their complexity, and the tokens deemed critical receive additional processing steps, while simpler tokens are processed with fewer resources, optimizing computational efficiency. 3) Thinking Step Encoding, introduces learnable position encodings for each step to distinguish between different processing steps, which helps the model organize and utilize information effectively during multi-step reasoning.

\paragraph{Reinforcement Learning (RL):}

Reinforcement Learning (RL) techniques have become prominent tools for optimizing Large Language Model (LLM) reasoning, specifically targeting test-time compute efficiency. 
A core motivation behind RL-based approaches is addressing the computational overhead resulting from unnecessarily verbose Chain-of-Thought (CoT) generations, which, while improving accuracy, substantially increase inference costs. 
Many works incorporate explicit reasoning-length penalties into the reward function to encourage brevity. These works mainly differ in the design of the reward function and RL algorithm. See Table~\ref{tab:rl_methods} for a comprehensive list.
For instance, Kimi k1.5~\citep{team2025kimi} includes a gradual warm-up strategy for the length penalty to mitigate slow early-phase training. 
Similarly, \citet{arora2025training} apply a tunable hyperparameter to trade off accuracy and brevity, showing token savings of $22\sim77\%$ while preserving up to 93\% of performance. 
Tencent's study~\citep{wang2025thoughtsplacethinking} introduces custom efficiency metrics and a self-training/RL strategy to prevent overthinking on trivial problems.
Meanwhile, \citet{yue2025vapoefficientreliablereinforcement} propose Length-Adaptive GAE, which dynamically adjusts the $\lambda$ parameter in $\lambda$-return used in Generalized Advantage Estimation (GAE) based on the sequence length, optimizing the balance between bias and variance for sequences of varying lengths, thus improving training stability and performance in long chain-of-thought tasks.

Other methods aim for more adaptive policies, enabling models to scale reasoning depth to task complexity. O1-Pruner~\citep{luo2025o1} applies a length-harmonizing fine-tuning method, integrating pre-sampled baseline estimations and accuracy-constrained RL fine-tuning to prune redundant reasoning systematically. 
Difficulty-Adaptive Slow-Thinking (DAST)~\citep{shen2025dast} introduces a difficulty-aware metric, Token Length Budget (TLB), guiding models through contrastive RL training to dynamically adjust reasoning lengths based on problem complexity. \citet{wang2025adaptivedeepreasoning} trains the LLM to toggle between short and long CoTs using reinforcement learning with an adaptive reward strategy that assesses problem complexity and provides corresponding rewards
Similarly, Think Smarter Not Harder~\citep{yu2025think} leverages Inference Budget-Constrained Policy Optimization (IBPO), framing reasoning as a constrained optimization problem to adaptively switch between short and long reasoning styles according to input complexity.

Improving RL exploration quality has also been a major theme to improve efficiency. Meta Reinforcement Fine-Tuning (MRT)~\citep{qu2025optimizing} moves beyond traditional binary correctness rewards by segmenting reasoning outputs into meaningful episodes and computing cumulative regret, providing finer feedback and improving generalization across compute budgets. 
Chain of Preference Optimization (CPO)~\citep{zhang2024chain} captures preference-aligned reasoning paths and integrates them into post-training objectives, approximating reasoning quality at lower inference costs. 
PANEL~\citep{li2025dancing} replaces scalar reward signals with natural language critiques of intermediate steps, yielding richer, more informative feedback for policy refinement.

\citet{wang2025adareasoneradaptivereasoning} introduce AdaReasoner, an LLM-agnostic plugin module that adaptively configures test-time compute parameters—such as system prompt, number of reasoning steps, and temperature—to enhance reasoning performance across various tasks. Unlike prior approaches, AdaReasoner frames this as a multi-armed bandit problem, where each arm represents a different configuration. The system is trained using a reinforcement learning framework, combining a factorized action space with a targeted exploration strategy and a pretrained reward model, enabling it to optimize reasoning configurations with only a few-shot guide. Experimental results demonstrate that AdaReasoner outperforms standard baselines, preserves out-of-distribution robustness, and yields gains on knowledge-intensive tasks through tailored inference parameters. However, the reward function does not incorporate a length penalty, leaving the efficiency of the approach unclear.

Recent works~\citep{fatemi2025concisereasoningreinforcementlearning,liu2025understanding,wu2025adaptivereasoningmodel,huang2025adactrladaptivecontrollable} critically evaluate existing RL optimizers, demonstrating that standard RL often introduces a length bias, favoring longer outputs due to reward signal propagation, even when they are suboptimal. These studies propose revised policy optimization methods such as Dr.GRPO~\citep{liu2025understanding} to explicitly control length explosion and stabilize verbosity during training, promoting concise reasoning as potentially more accuracy-enhancing. Some methods~\citep{jiang2025thinkneedlarge,huang2025adactrladaptivecontrollable, wu2025adaptivereasoningmodel} such as AdaCtrl~\cite{huang2025adactrladaptivecontrollable} include a finetuning stage that enables users to trigger different reasoning strategies/budgets through pre-defined tags (e.g., ``[Easy]''). This is followed by a RL training stage to refine the LLM's ability to adapt the reasoning strategy based on question difficulty. The combination of the two training stages gives such methods the ability to adjust the response based on either the \textit{user-specified budget (L1)} or the \textit{model’s own difficulty-awareness (L2)}.

\begin{tcolorbox}[
  enhanced,
  colback=white,
  colframe=black,
  boxrule=1.0pt,
  fonttitle=\bfseries,
  title=Remarks on RL for Eliciting Efficiency,
  attach boxed title to top left={yshift=-2mm, xshift=2mm},
  boxed title style={
    colback=MidnightBlue,
    colframe=black,
    boxrule=0pt,
    fontupper=\color{white}\bfseries
  }
]
RL enables both explicit length control and dynamic adaptation of reasoning depth in LLMs. Through concise reward shaping, budget-aware inference, or richer feedback mechanisms, RL provides a flexible framework for aligning computational resources effectively with reasoning requirements. However, while RL can provide better out-of-distribution generalization compared to SFT, it requires significantly more training and has limited effectiveness when applied to extremely underfit or overfit model checkpoints~\citep{chu2025sftmemorizesRLgeneralizes}.
\end{tcolorbox}

\renewcommand{\arraystretch}{1.5}

\begin{table}[!ht]
\centering
\caption{Comparison of different length reward-based RL methods. The last two columns represent the data used for training and/or evaluation.}
\resizebox{\textwidth}{!}{%
\begin{tabular}{p{5cm}p{2.5cm}lll}
\toprule
\large \textbf{Method} & \large \textbf{RL} & \large \textbf{Length Constraint Reward} & \large \textbf{STEM} & \large \textbf{General}  \\ \midrule
O1-Pruner~\citep{luo2025o1}  & PPO & $\mathbb{E}_{x\sim D}[\mathbb{E}_{\pi_\theta,\pi_{ref}}[\frac{L(y_{ref})}{L(y_{pred})}] - 1] + \lambda(A(x, y) - A(x, y'))$ & \checkmark  &
\\ \hline
Demystifying~\citep{yeo2025demystifyingLongCoT} & PPO & $\begin{cases} r_{c0}+0.5(r_{cL}-r_{c0})(1+\cos(\frac{\pi L(y_{pred})}{L_{max}})), & \text{if correct} \\ r_{c0}+0.5(r_{wL}-r_{w0})(1+\cos(\frac{\pi L(y_{pred})}{L_{max}})), & \text{if wrong} \\ r_e, & \text{if } L(y_{pred})=L_{max} \end{cases}$ & \checkmark & \checkmark
\\ \hline
L1~\citep{aggarwal2025l1controllinglongreasoning}  & GRPO & $r(y,y_{GT},n_{GT})=I(y=y_{GT})-\alpha \cdot |n_{GT}-n_y|$ & \checkmark & \checkmark
\\ \hline
DAST~\citep{shen2025dast}  & SimPO & Trained with constructed length preference data & \checkmark &  
\\ \hline
Training~\citep{arora2025training} & Policy Gradient & $\mathbb{E}_{x\sim D}[1\{y_{pred}=y_{GT}\}(1-\alpha f(L(y_{pred})))]$ & \checkmark & 
\\ \hline
Kimi k1.5~\citep{team2025kimi}  & Online Policy Mirror Decent & \makecell[l]{$\text{len\_reward}(i)=\begin{cases}\lambda, & \text{if } r(x,y_i,y^*)=1 \\ \min(0, \lambda), & \text{if } r(x,y_i,y^*)=0\end{cases}$ \\ $\lambda=0.5-\frac{\text{len}(i)-\text{min\_len}}{\text{max\_len}-\text{min\_len}}$} & \checkmark & \checkmark
\\ \hline
VAPO~\citep{yue2025vapoefficientreliablereinforcement} & Value-based PPO & \makecell[l]{\text{Length-adaptive λ in GAE:} $\lambda_{policy} = 1 - \frac{1}{\alpha \cdot l}$,\\ \text{ where } $\alpha = 0.05$} & \checkmark &  
\\ \hline
MRT~\citep{qu2025optimizing} & GRPO / STaR & \makecell[l]{$\ell_{\text{MRT}} = \ell_{\text{FT}} + \alpha \sum_j r^{\mu}_{\text{prg}}(z_j; c)$, \\ where $r^{\mu}_{\text{prg}}(z_j; c) = J_r(\mu(\cdot|z_j, c)) - J_r(\mu(\cdot|c))$} & \checkmark & 
\\ \hline
Dr. GRPO~\citep{liu2025understanding} & GRPO & \makecell[l]{$A_t = R(q, o) - \text{mean}(\{R(q, o_1), ..., R(q, o_G)\})$ \\ (no length/std normalization)} & \checkmark & 
\\ \hline
Concise Reasoning \citep{fatemi2025concisereasoningreinforcementlearning} & PPO & $\begin{cases} +1, & \text{if correct boxed answer} \\ -0.5, & \text{if incorrect boxed answer} \\ -1, & \text{if no boxed answer} \end{cases}$ (with \(\lambda < 1\) in GAE) & \checkmark & \\ \hline
Adaptive Deep Reasoning \citep{wang2025adaptivedeepreasoning} & GRPO & $\begin{cases}
			-1.0 & \text{Incorrect Answer} \\[6pt]
			\begin{aligned}
				\text{Case } \alpha > \theta: &\quad
				\begin{cases} 
					+1.5 & \text{(Short CoT)} \\ 
					+1.0 & \text{(Long CoT)}
				\end{cases} \\[6pt]
				\text{Case } \alpha \leq \theta: &\quad
				\begin{cases} 
					+1.0 & \text{(Short CoT)} \\ 
					+1.5 & \text{(Long CoT)}
				\end{cases}
			\end{aligned}
			& \text{Correct Answer}
		\end{cases}$  & \checkmark & 
\\ \hline
ARM~\citep{wu2025adaptivereasoningmodel} & GRPO & \makecell[l]{$r = 1\{y_{pred}=y_{GT}\} \alpha_i(t)$ \\ \text{Diversity Scaling:} $\alpha_i(t) = \frac{G}{F(o_i)}  decay_i(t)$} & \checkmark & \checkmark \\ \bottomrule
\end{tabular}%
}
\label{tab:rl_methods}
\end{table}

\paragraph{Model Merging:} While SFT and RL are popular ways to tackle the overthinking problem, they require additional training and are computationally expensive. Furthermore, the training could be unstable, and has non-trivial engineering considerations. \citet{wu2025unlockingefficientlong} explore the possibility of merging the quick thinking model with its reasoning counterpart as another method to address the issue. Here, they find that activation based merging methods work the best, resulting in a 50-55\% reduction in token length compared to the slow thinking model, while maintaining performance, or in some cases, even achieving higher scores. While these methods were promising for 1.5B and 7B models, merging of 14B and 32B models has disappointing results, requiring either a sacrifice of accuracy or leading to only small (\textless10\%) reductions in token length.

\subsection{Parallel}\label{sec:l2_parallel}

\paragraph{Prompting:} \citet{aggarwal2023adaptive} observe that  for many standard benchmarks and LLMs, the accuracy of \emph{self-consistency} does not improve when increasing the number of CoTs after a few samples.
In order to save the redundant computational budget, they introduce a stopping criterion for self-consistency  based on the confidence in the most frequent answer over
other answers. This method iteratively samples CoTs until a model-agnostic and lightweight Stopping Criterion is satisfied; it then reports the majority answer. ~\citet{wang2024make} propose a similar approach, grouping problems by estimated difficulty and then allocating the number of CoTs accordingly. A practical limitation of this method, which may inhibit streaming applications, is that it requires the knowledge of all the questions during the difficulty ranking phase. A more general approach proposed by~\citet{li2024escape} alleviates this issue by introducing early stopping to self-consistency runs, halting computation when the responses have sufficiently converged. This is framed as a hypothesis test with the null hypothesis that the current plurality answer is not the mode of the underlying distribution driving the self consistency samples. While this approach is more versatile and achieves good empirical performance, it comes with the drawback that the statistical guarantees of the hypothesis test are valid only for larger self-consistency runs where the normality assumptions made in the hypothesis test are valid. \citet{zhu2024path_consistency} also observe that vanilla self-consistency wastes a significant percentage of tokens on generating CoTs that lead to incorrect solutions. In order to save inference cost, the authors propose Path-consistency, which uses partial reasoning steps from previously generated CoTs for subsequent generations. The method begins by generating a small number of reasoning paths. Then, using a confidence assessment metric, the current optimal reasoning path is identified and an initial short prefix (e.g., the first few steps of the current optimal reasoning path) is extracted. This initial prefix is then added to the prompt for sampling the next batch of branches. This procedure is repeated iteratively, progressively extending the prefix length until the sampled answers satisfy a  suitable stopping criterion (e.g., the Beta criterion proposed by~\citet{aggarwal2023adaptive}).

More recent work combines generating multiple CoTs with the utilization of reward models or verifiers to score them. \citet{snell2024scaling} propose selection of the answer based on the highest-scored CoT; this approach is termed \emph{Best-of-N} (BoN) and is shown to outperform self-consistency on challenging benchmarks.
However, \citet{wang2025self-estimating} note that sampling multiple reasoning paths in their entirety is computationally wasteful and propose a method to focus on  promising CoTs. This technique, termed Self-Truncation (ST)-BoN, identifies an early estimation point in time, where candidate reasoning paths begin to diverge from each other. It then performs continuous evaluation within a buffer window to select the most promising candidate to continue generation, while the others are truncated, significantly reducing computational cost.
\citet{fu2024certainindex} introduce an LLM serving system which can adapt to the scaling behavior of various parallel sampling algorithms and the varying
difficulty of queries. This leads to efficient resource allocation and reduced latency.
Specifically, the scheduler system dynamically allocates compute to different reasoning queries while they are run and estimates the individual trade-offs between allocating more compute and progress towards a final answer. This work introduces a reasoning progress metric, termed certainindex, which is based on semantic entropy and is used to measure uncertainty during answer generation for a variety of TTC algorithms (e.g. BoN, SC, and o1-like models). The proposed certainindex variants show positive correlation with reasoning difficulty in contrast to other heuristic notions of LLM reasoning certainty such as CoT length and mean normalized log-probability. However, computing some variants of this metric might be difficult for o1-like models.

\paragraph{Supervised Finetuning:} \citet{wan2024rasc} introduce a novel method, termed \emph{reasoning aware self-consistency} (RASC),  for increasing the sampling efficiency of self-consistency via early stopping.  RASC extracts a set of heuristic features from each CoT that measure its reasoning quality (e.g., length, model confidence, and diversity) and trains a scoring function that predicts correctness from these features.
The CoTs surpassing a threshold score are included in a set of high-quality feasible CoTs. 
Subsequently, a score-weighted majority voting on this feasible set is carried out to choose the final answer. \citet{damani2025learning} propose a method that can be applied in different test-time compute settings. For example, for the Best-of-N~\citep{snell2024scaling} approach, they estimate how many CoTs should be sampled before applying the reward model to rank them. 
For routing between two LLMs, they learn the probability of obtaining a higher reward from the computationally expensive but stronger LLM in comparison to a weaker but cost-effective LLM.  In both cases, queries are assigned to the appropriate decoding strategy based on their estimated difficulties by solving an online/offline budget allocation problem. \citet{yu2025think} introduce a fine-tuned constrained optimization objective to learn the difficulty of queries without an external verifier and allocate inference budgets for harder questions. While this approach allows the model to operate efficiently without external methods, it is cumbersome to train as it requires solving an NP-hard integer linear programming problem~\citep{vazirani2010ApproximationAlgorithms}.

\paragraph{Reinforcement Learning:} \citet{pan2025learningadaptiveparallel} were among the first to apply reinforcement learning for L2 parallel methods. Instead of imposing fixed reasoning structures or relying on external strategies like depth-first search (DFS), the authors train an LLM to dynamically spawn independent child processes via a \texttt{spawn()} operation. These child threads explore distinct reasoning paths in parallel and return their results to the parent thread using a \texttt{join()} operation. The parent then continues decoding, conditioned on the aggregated outcomes. The model is fine-tuned end-to-end using reinforcement learning to optimize task performance without predefined reasoning templates. Experiments on the Countdown game~\citep{yao2023treeofthought} show improved token efficiency, though generalization to broader reasoning tasks (e.g., mathematics) remains an open question.

\section{Discussion}\label{sec:discussion}

\paragraph{Prompting or Fine-Tuning? Selecting the Right Tool to Elicit Efficient Reasoning:} All methods discussed in this survey aim to improve the efficiency of test-time compute, but which approach should a user choose? For users with limited training budgets, prompting-based methods are the most accessible. Techniques such as self-consistency~\citep{wang2023selfconsistency} and BoN~\citep{cobbe2021trainingverifierssolve} offer L1 controllability via the number of sampled responses. However, these methods may still generate arbitrarily long outputs for each response, which can lead to inefficiencies. For more fine-grained control, users can combine them with L1 sequential strategies such as s1~\citep{muennighoff2025s1simpletest} or CCot~\citep{renze2024BenefitsConciseChain}, which focus on reducing token length without sacrificing performance.
For users who can afford training, finetuning-based methods present a rich set of options. These include reinforced supervised fine-tuning~\citep{yang2025thinkingoptimalscaling, munkhbat2025selftrainingelicits} and reinforcement learning approaches~\citep{aggarwal2025l1controllinglongreasoning, team2025kimi}, which can imbue models with more efficient reasoning habits. A compelling direction for future research is to systematically compare fine-tuning methods in terms of how they improve the efficiency of baseline TTC strategies.

\paragraph{Towards Hybrid Fast-Slow LLMs:} While large reasoning models (e.g., o1-series~\citep{openai2024openaio1card, openai2025o3}) have had immense success in complex reasoning and research-intensive tasks (e.g., math, science, etc), vanilla fast LLMs (e.g., GPT-4o~\citep{openai2024gpt4o}, DeepSeek-v3~\citep{deepseek2025deepseekv3}) remain competitive in other language understanding tasks such as creative writing, translation, and text retrieval. This is largely due to the use of reinforcement learning to enhance reasoning, which requires well-defined rules and verifiable correctness criteria to enable effective reward modeling. Extending these techniques to broader reasoning domains presents significant challenges. There is often limited training data for RL due to the difficulty of defining verifiable rewards, and it is hard to ensure generalization across diverse tasks. Therefore, an interesting direction for future work would be designing hybrid models that incorporate both intuitive (fast) and complex (slow) reasoning capabilities. Furthermore, new LLM architectures such as diffusion models~\citep{nie2025largeLanguageDiffusion,sahoo2024simplEffectiveMaskeded} may provide a better framework for incorporating multiple reasoning modalities by enabling iterative refinement in a way that complements both fast and slow inference stages.

\paragraph{Controllable and Efficient LLM Test-time Compute Beyond Language Models:} Complex reasoning capabilities, such as questioning and reflection, have been extended to multi-modal large language models (MLLMs)~\citep{yang2025r1, zhang2025r1, feng2025video, zhou2025r1}. However, very few works study efficient test-time compute (TTC) in the context of MLLM reasoning. 
For example, Vision-R1~\citep{huang2025vision} introduces a training pipeline combining SFT and RL to promote complex reasoning in MLLMs. To improve token efficiency, it introduces Progressive Thinking Suppression Training (PTST), which initially constrains the reasoning length but then gradually relaxes this constraint over the course of training. This adaptive approach outperforms fixed length chain-of-thought (CoT) training under equivalent CoT budgets. LLaVA-CoT~\citep{xu2025llavacotletvisionlanguage} adopts a parallel reasoning strategy equipped with stage-level beam search. The model is finetuned via SFT using structured reasoning data annotated with four distinct thinking stages. During inference, the top response from $N$ candidates at each stage is propagated to the next, enabling more efficient reasoning. This stage-level beam search surpasses both best-of-N and sentence-level beam search under similar token constraints. 

For the L2 paradigm, FAST~\citep{xiao2025fast} introduces a fast-slow thinking framework that dynamically adjusts reasoning length based on the characteristics of each question. Specifically, it defines two metrics: question \textit{difficulty}, measured by pass$@$k, and image \textit{complexity}, based on the complexity of the image texture and semantics. The training data are categorized into \texttt{Hard} and \texttt{Easy} subsets based on the \textit{difficulty} metric, with \texttt{Hard} examples used in the early training epochs to encourage slow, deliberate reasoning, and \texttt{Easy} examples used later to promote fast thinking. During GRPO training, a complexity-based thinking reward and difficulty-aware KL regularization are employed to balance fast and slow thinking while dynamically controlling exploration. The method is evaluated on MathVista and achieves comparable performance with reduced average reasoning length. Similarly, Skywork R1V~\citep{peng2025skyworkr1vpioneeringmultimodal} proposes an Adaptive-Length Chain-of-Thought Distillation (AL-CoTD) method to dynamically adjust reasoning chain lengths based on task complexity, preventing overthinking and excessive computation. VLM-R1~\citep{shen2025vlm} uses RL finetuning for open-vocabulary object detection (OVD) task and introduces a penalty factor for accuracy reward to reduce redundant prediction from VLMs. The penalty factor $s_{ovd}$ is defined as $s_{ovd} = \min(1, L_{gt} / L_{pred})$, where $L_{gt}$ is the number of ground truth object labels and $L_{pred}$ is the number of predicted objects. The accuracy reward is thus computed as $R_{acc}^{ovd}=s_{pen} \cdot mAP$, where $mAP$ is the mean average precision metric. Experimental results show that the proposed reward with this penalty factor significantly reduces the completion length while achieving better performance than the naive $mAP$ reward.

\paragraph{Applications:} Efficiency is critical for real-world applications. Companies often deploy multiple model sizes to serve both latency-sensitive and compute-rich use cases~\citep{openai2025o3, alibaba2025qwen3}. L1-controllable methods offer a promising path toward “one-size-fits-all” models that dynamically adjust compute usage based on user constraints.
Additionally, L1 and L2 methods are especially relevant for interactive AI agents (e.g., Deep Research~\citep{openai2025deepresearch}) that integrate external tools such as search engines. These agents frequently operate under tight latency requirements while maintaining high response quality.
In multi-modal domains such as autonomous driving~\citep{liao2025cotdriveefficientmotion,yang2024llmdrivesurvey}, robotics~\citep{Zawalski2024RoboticControlEmbodied, intelligence2025pi05visionlanguage}, and healthcare~\citep{alSaad2024MultimodalLargeLanguage}, LLMs have demonstrated impressive reasoning capabilities. However, current efficiency efforts in this space primarily focus on model compression techniques—quantization~\citep{xie2024advancingMultimodalLarge}, distillation~\citep{hegde2025distillingmultimodallarge}, and pruning~\citep{Ye20205fitprunemultimodal}. Future work should explore reasoning-level optimization strategies, such as controlling the length of generated chains-of-thought (CoTs) and minimizing redundant reasoning steps, to better support real-time multi-modal applications.

\section{Conclusion}\label{sec:conclusion}

This survey presented a comprehensive review of test-time compute (TTC) strategies in large language models (LLMs), introducing a novel two-tiered taxonomy that distinguishes between controllable (L1) and adaptive (L2) efficiency. Through extensive benchmarking, we identified critical inefficiencies in current models that rely on fixed compute budgets, underscoring the need for dynamic, task-aware inference. A promising direction for future work is the development of hybrid fast–slow thinking LLMs that can flexibly allocate reasoning effort based on task complexity. Beyond language tasks, TTC methods offer exciting opportunities for multi-modal foundation models, including vision-language and embodied AI, where adaptive compute could substantially improve performance in real-world, interactive settings. Advancing toward models that unify control and adaptiveness across modalities may unlock the next generation of efficient, scalable, and context-aware AI.

\bibliographystyle{unsrtnat}
\bibliography{draft}

\appendix
\section{Experimental Results}\label{app:exp_setup}

We describe the experimental setup used to generate the results in Figures \ref{fig:claude_controllability} and \ref{fig:model_perf_acc_token_count}. The detailed per-dataset accuracy results can be found in Table~\ref{tab:per_dataset_perf}.

\paragraph{Experimental Settings:} For inference, we used a temperature of 0.7 for all models. The Qwen3 series and DeepSeek R1-Distilled Qwen-7B were deployed locally using vLLM, with the maximum token length set to 32,768. Other commercial models were accessed via their respective APIs. For Gemini, the thinking "on" setting corresponds to a thinking budget of 24,576 tokens, while "off" sets the budget to zero. For Claude, "on" uses a budget of 21,332 tokens, and "off" is set to zero. The GPT-series and DeepSeek-R1 models used default inference settings.

For evaluation, all test sets report pass@1, except for AIME due to its small size (30 questions each for 2024 and 2025). For AIME, we performed eight independent inference and evaluation runs, reporting the average accuracy. All model outputs were evaluated using the LLM-as-judge framework with GPT-4o as the judge, using a temperature of 0.3 and averaging over three runs for each test set.

\renewcommand{\arraystretch}{2.5}
\begin{table}[H]
\centering
\resizebox{\textwidth}{!}{
\begin{tabular}{l l
  *{4}{>{\centering\arraybackslash}p{1.5cm}}  
  *{3}{>{\centering\arraybackslash}p{1.5cm}}  
  *{3}{>{\centering\arraybackslash}p{1.5cm}}  
}
\toprule
\textbf{Model} & \textbf{Reasoning Effort}
& \multicolumn{4}{c}{\textbf{Math}} 
& \multicolumn{3}{c}{\textbf{STEM (MMLU Pro)}} 
& \multicolumn{3}{c}{\textbf{Non-STEM (MMLU Pro)}} \\
\cmidrule(lr){3-6} \cmidrule(lr){7-9} \cmidrule(lr){10-12}
& 
& \textbf{ASDiv} & \textbf{GSM-8K} & \textbf{MATH-500} & \textbf{AIME-24\&25}
& \textbf{Phy.} & \textbf{Chem.} & \textbf{Bio.}
& \textbf{His.} & \textbf{Law} & \textbf{Bus.} \\
& 
& Acc / Avg. Tok. & Acc / Avg. Tok. & Acc / Avg. Tok. & Acc / Avg. Tok.
& Acc / Avg. Tok. & Acc / Avg. Tok. & Acc / Avg. Tok.
& Acc / Avg. Tok. & Acc / Avg. Tok. & Acc / Avg. Tok. \\
\midrule

\multirow{3}{*}{\textbf{GPT-o1}} 
& low  &\makecell{99.34\% /\\ 113.39} &\makecell{97.04\% /\\ 311.94} &\makecell{95.2\% /\\ 1436.01} &\makecell{71.67\% /\\ 7973.68} &\makecell{90.47\% /\\ 1434.17} &\makecell{94\% /\\ 891.45} &\makecell{82.68\% /\\ 930.25} &\makecell{80.58\% /\\ 975.35} &\makecell{82.92\% /\\ 1263.30} &\makecell{88.42\% /\\ 1018.61} \\
& med  &\makecell{99.34\% /\\ 162.85} &\makecell{96.82\% /\\ 393.56} &\makecell{97\% /\\ 1861.78} &\makecell{75\% /\\ 10769.83} &\makecell{91.84\% /\\ 2046.27} &\makecell{89.55\% /\\ 2314.79} &\makecell{94\% /\\ 1268.73} &\makecell{82.15\% /\\ 1365.79} &\makecell{83.47\% /\\ 1925.61} &\makecell{89.74\% /\\ 1485.68} \\
& high &\makecell{99.26\% /\\ 214.99} &\makecell{96.89\% /\\ 508.59} &\makecell{98.2\% /\\ 2435.02} &\makecell{83.33\% /\\ 14067.28} &\makecell{92.07\% /\\ 2941.90} &\makecell{89.73\% /\\ 3312.99} &\makecell{95.54\% /\\ 1706.52} &\makecell{83.2\% /\\ 1914.50} &\makecell{84.11\% /\\ 2732.09} &\makecell{91.41\% /\\ 2237.7} \\
\midrule
\multirow{2}{*}{\textbf{Gemini 2.5 Flash}} 
& Off  
& \makecell{99.18\% /\\ 169.01} 
& \makecell{95.38\% /\\ 334.28} 
& \makecell{95.20\% /\\ 1659.69} 
& \makecell{63.30\% /\\ 13263.97} 
& \makecell{88.84\% /\\ 1984.18} 
& \makecell{88.52\% /\\ 2150.08} 
& \makecell{93.86\% /\\ 929.13} 
& \makecell{78.74\% /\\ 874.15} 
& \makecell{72.42\% /\\ 1278.65} 
& \makecell{88.59\% /\\ 1096.75} \\
& On 
& \makecell{99.18\% /\\ 543.21} 
& \makecell{96.01\% /\\ 1349.20} 
& \makecell{98.60\% /\\ 6925.86} 
& \makecell{76.67\% /\\ 23394.02} 
& \makecell{91.15\% /\\ 7158.66} 
& \makecell{90.90\% /\\ 6613.30} 
& \makecell{95.11\% /\\ 2407.89} 
& \makecell{83.46\% /\\ 2220.52} 
& \makecell{80.81\% /\\ 3544.08} 
& \makecell{89.61\% /\\ 3732.67} \\
\midrule
\textbf{DeepSeek-R1}
& 671B  
& \makecell{99.18\% /\\ 907.78} 
& \makecell{97.81\% /\\ 1391.62} 
& \makecell{96.2\% /\\ 3021.56} 
& \makecell{63.33\% /\\ 6377.80} 
& \makecell{88.84\% /\\ 3732.29} 
& \makecell{89.83\% /\\ 3820.27} 
& \makecell{94\% /\\ 2069.01} 
& \makecell{81.63\% /\\ 1947.07} 
& \makecell{79.93\% /\\ 2828.85} 
& \makecell{89.1\% /\\ 2848.48} \\
\midrule
\multirow{2}{*}{\makecell{\textbf{DeepSeek-R1}\\\textbf{Distilled Qwen-2.5}}} 
& 7B
& \makecell{94.17\% /\\ 1168.25} 
& \makecell{89.69\% /\\ 1753.03} 
& \makecell{89.60\% /\\ 3696.75} 
& \makecell{43.33\% /\\ 11632.27} 
& \makecell{70.90\% /\\ 6759.32} 
& \makecell{67.76\% /\\ 7559.58} 
& \makecell{59.97\% /\\ 2939.43} 
& \makecell{26.25\% /\\ 2573.97} 
& \makecell{30.70\% /\\ 3367.01} 
& \makecell{64.89\% /\\ 5528.27} \\
& 32B  
& \makecell{97.21\% /\\ 1272.36} 
& \makecell{96.44\% /\\ 1467.28} 
& \makecell{93.80\% /\\ 3229.25} 
& \makecell{61.67\% /\\ 9945.48} 
& \makecell{83.99\% /\\ 4695.19} 
& \makecell{79.51\% /\\ 5594.29} 
& \makecell{84.38\% /\\ 2603.93} 
& \makecell{64.30\% /\\ 1854.75} 
& \makecell{59.58\% /\\ 2151.73} 
& \makecell{78.71\% /\\ 4426.33} \\
\midrule
\multirow{2}{*}{\textbf{Claude3.7}} 
& Off 
& \makecell{98.84\% /\\ 162.79} 
& \makecell{96.51\% /\\ 289.01} 
& \makecell{81.58\% /\\ 571.12} 
& \makecell{18.96\% /\\ 965.95} 
& \makecell{84.50\% /\\ 579.21} 
& \makecell{84.40\% /\\ 686.59} 
& \makecell{93.44\% /\\ 466.51} 
& \makecell{79.42\% /\\ 502.06} 
& \makecell{75.73\% /\\ 611.71} 
& \makecell{86.48\% /\\ 435.41} \\
& On 
& \makecell{99.26\% /\\ 504.47} 
& \makecell{97.11\% /\\ 1380.68} 
& \makecell{94.18\% /\\ 6439.13} 
& \makecell{47.17\% /\\ 16403.36} 
& \makecell{91.16\% /\\ 6523.27} 
& \makecell{90.81\% /\\ 7032.57} 
& \makecell{95.52\% /\\ 2855.83} 
& \makecell{78.89\% /\\ 2175.04} 
& \makecell{81.03\% /\\ 4444.63} 
& \makecell{90.97\% /\\ 4159.25} \\
\midrule
\multirow{2}{*}{\textbf{Qwen3-4B}} 
& Off 
& \makecell{98.93\% /\\ 129.53} 
& \makecell{92.65\% /\\ 317.98} 
& \makecell{87.20\% /\\ 993.45} 
& \makecell{21.67\% /\\ 4243.97} 
& \makecell{72.06\% /\\ 1109.70} 
& \makecell{71.73\% /\\ 1262.75} 
& \makecell{77.41\% /\\ 783.58} 
& \makecell{52.49\% /\\ 613.86} 
& \makecell{45.69\% /\\ 879.43} 
& \makecell{71.86\% /\\ 914.48} \\
& On  
& \makecell{99.51\% /\\ 1061.72} 
& \makecell{96.51\% /\\ 2374.83} 
& \makecell{97.20\% /\\ 5356.69} 
& \makecell{80.00\% /\\ 15526.07} 
& \makecell{87.53\% /\\ 6161.14} 
& \makecell{86.13\% /\\ 7020.12} 
& \makecell{86.61\% /\\ 2597.65} 
& \makecell{64.30\% /\\ 2153.03} 
& \makecell{52.95\% /\\ 3201.27} 
& \makecell{84.41\% /\\ 4749.70} \\
\midrule
\multirow{2}{*}{\textbf{Qwen3-8B}} 
& Off 
& \makecell{99.43\% /\\ 133.19} 
& \makecell{93.56\% /\\ 323.96} 
& \makecell{85.40\% /\\ 1225.19} 
& \makecell{26.67\% /\\ 4281.78} 
& \makecell{76.75\% /\\ 1071.34} 
& \makecell{76.86\% /\\ 1220.60} 
& \makecell{83.26\% /\\ 959.41} 
& \makecell{65.09\% /\\ 837.71} 
& \makecell{53.59\% /\\ 1205.43} 
& \makecell{75.54\% /\\ 844.85} \\
& On  
& \makecell{99.59\% /\\ 1319.03} 
& \makecell{97.35\% /\\ 2465.33} 
& \makecell{98.00\% /\\ 5515.70} 
& \makecell{80.00\% /\\ 16006.77} 
& \makecell{89.30\% /\\ 6423.94} 
& \makecell{88.69\% /\\ 7292.11} 
& \makecell{92.05\% /\\ 2810.70} 
& \makecell{68.77\% /\\ 2362.54} 
& \makecell{63.40\% /\\ 3924.67} 
& \makecell{86.82\% /\\ 4988.30} \\
\midrule
\multirow{2}{*}{\textbf{Qwen3-14B}} 
& Off 
& \makecell{99.26\% /\\ 129.79} 
& \makecell{95.53\% /\\ 318.41} 
& \makecell{88.60\% /\\ 939.83} 
& \makecell{35.00\% /\\ 3427.50} 
& \makecell{80.60\% /\\ 926.40} 
& \makecell{79.86\% /\\ 1139.15} 
& \makecell{86.19\% /\\ 712.84} 
& \makecell{66.40\% /\\ 631.87} 
& \makecell{57.67\% /\\ 992.92} 
& \makecell{80.35\% /\\ 723.54} \\
& On  
& \makecell{99.51\% /\\ 1007.24} 
& \makecell{97.80\% /\\ 1980.85} 
& \makecell{98.20\% /\\ 4805.85} 
& \makecell{86.44\% /\\ 13779.63} 
& \makecell{91.53\% /\\ 5718.72} 
& \makecell{90.19\% /\\ 6553.33} 
& \makecell{92.61\% /\\ 2415.03} 
& \makecell{75.07\% /\\ 2019.22} 
& \makecell{68.39\% /\\ 3126.68} 
& \makecell{88.97\% /\\ 4429.67} \\
\midrule
\textbf{DeepSeek-V3} & -- 
& \makecell{99.01\% /\\ 245.81} 
& \makecell{96.97\% /\\ 387.78} 
& \makecell{95.60\% /\\ 1405.44} 
& \makecell{48.33\% /\\ 3660.57} 
& \makecell{90.53\% /\\ 1123.91} 
& \makecell{87.19\% /\\ 1095.80} 
& \makecell{94.56\% /\\ 703.36} 
& \makecell{75.85\% /\\ 589.80} 
& \makecell{75.11\% /\\ 964.87} 
& \makecell{89.73\% /\\ 850.97} \\
\textbf{GPT-4o} & -- 
& \makecell{99.18\% /\\ 123.02} 
& \makecell{96.06\% /\\ 278.07} 
& \makecell{78.40\% /\\ 590.60} 
& \makecell{15.00\% /\\ 1222.23} 
& \makecell{77.83\% /\\ 575.38} 
& \makecell{75.18\% /\\ 677.17} 
& \makecell{91.91\% /\\ 322.47} 
& \makecell{78.22\% /\\ 337.67} 
& \makecell{71.84\% /\\ 440.78} 
& \makecell{79.85\% /\\ 404.64} \\
\bottomrule
\end{tabular}
}
\caption{Per-dataset Pass@1 Accuracy and Average Token Count for Each Model}
\label{tab:per_dataset_perf}
\end{table}

\end{document}

%% file: Structure.tex
\usepackage{amsmath,amsfonts,stmaryrd,amssymb,mathtools} 
\usepackage{cancel}

\usepackage{booktabs}

\usepackage{enumerate} 

\usepackage[framemethod=tikz]{mdframed} 

\usepackage{hyperref}

\usepackage{algorithm}
\usepackage{algorithmicx}
\usepackage{algpseudocode}

\usepackage{pdfpages}

\usepackage{pifont}

\usepackage[capitalise]{cleveref}
\usepackage{natbib}
\PassOptionsToPackage{sort, numbers, compress}{natbib}

\usepackage{listings} 
\usepackage{geometry} 

\usepackage{setspace}
\usepackage{lipsum}
\usepackage{etoolbox}
\AtBeginEnvironment{quote}{\singlespace\vspace{-\topsep}\small}
\AtEndEnvironment{quote}{\vspace{-\topsep}\endsinglespace}

\usepackage[utf8]{inputenc} 
\usepackage[T1]{fontenc} 

\usepackage{XCharter} 

\usepackage[LGR,T1]{fontenc} 
\usepackage[utf8]{inputenc}

\mdfdefinestyle{commandline}{
	leftmargin=10pt,
	rightmargin=10pt,
	innerleftmargin=15pt,
	middlelinecolor=black!50!white,
	middlelinewidth=2pt,
	frametitlerule=false,
	backgroundcolor=black!5!white,
	frametitle={Command Line},
	frametitlefont={\normalfont\sffamily\color{white}\hspace{-1em}},
	frametitlebackgroundcolor=black!50!white,
	nobreak,
}

\mdfdefinestyle{file}{
	innertopmargin=1.6\baselineskip,
	innerbottommargin=0.8\baselineskip,
	topline=false, bottomline=false,
	leftline=false, rightline=false,
	leftmargin=2cm,
	rightmargin=2cm,
	singleextra={%
		\draw[fill=black!10!white](P)++(0,-1.2em)rectangle(P-|O);
		\node[anchor=north west]
		at(P-|O){\ttfamily\mdfilename};
		\def\l{3em}
		\draw(O-|P)++(-\l,0)--++(\l,\l)--(P)--(P-|O)--(O)--cycle;
		\draw(O-|P)++(-\l,0)--++(0,\l)--++(\l,0);
	},
	nobreak,
}

\mdfdefinestyle{question}{
	innertopmargin=1.2\baselineskip,
	innerbottommargin=0.8\baselineskip,
	roundcorner=5pt,
	nobreak,
	singleextra={%
		\draw(P-|O)node[xshift=1em,anchor=west,fill=white,draw,rounded corners=5pt]{%
		Question \theQuestion\questionTitle};
	},
}

\newcounter{Question} 

\mdfdefinestyle{warning}{
	topline=false, bottomline=false,
	leftline=false, rightline=false,
	nobreak,
	singleextra={%
		\draw(P-|O)++(-0.5em,0)node(tmp1){};
		\draw(P-|O)++(0.5em,0)node(tmp2){};
		\fill[black,rotate around={45:(P-|O)}](tmp1)rectangle(tmp2);
		\node at(P-|O){\color{white}\scriptsize\bf !};
		\draw[very thick](P-|O)++(0,-1em)--(O);
	}
}

\mdfdefinestyle{info}{%
	topline=false, bottomline=false,
	leftline=false, rightline=false,
	nobreak,
	singleextra={%
		\fill[black](P-|O)circle[radius=0.4em];
		\node at(P-|O){\color{white}\scriptsize\bf i};
		\draw[very thick](P-|O)++(0,-0.8em)--(O);
	}
}
